\documentclass[a4paper,11pt]{article}
\usepackage[a4paper,margin=1.2in]{geometry}
\usepackage[numbers,sort]{natbib}
\usepackage{algorithm,algpseudocode}
\algnewcommand{\Result}[1]{\State \textbf{Result:} #1}
\usepackage{array}
\usepackage{textcomp}
\usepackage{stfloats}
\usepackage{url}
\usepackage{verbatim}
\usepackage{graphicx}
\usepackage{balance}
\usepackage{amssymb}
\usepackage{ifthen}
\usepackage[utf8]{inputenc}
\usepackage{latexsym, dsfont}
\usepackage{amsmath,amsthm}
\usepackage{booktabs}
\usepackage{siunitx}
\usepackage{graphicx}
\usepackage{caption}
\usepackage{hyperref}   
\usepackage[capitalize]{cleveref}
\usepackage{todonotes}
\usepackage[caption=false,font=footnotesize]{subfig}
\usepackage{mathtools}
\usepackage{tikz}
\newboolean{PGFPlots}
\setboolean{PGFPlots}{false}
\usetikzlibrary{circuits.logic.US}
\pgfdeclarelayer{background}
\pgfsetlayers{background,main}
\usetikzlibrary{arrows,shapes,decorations,decorations.pathreplacing,decorations.pathmorphing,intersections,calc,fit}
\ifthenelse{\boolean{PGFPlots}}{
\usepackage{pgfplots}
\usepgfplotslibrary{units}
\usetikzlibrary{external}
\tikzexternalize
\tikzset{external/only named=true}
}{}

\newcommand{\IR}{\mathds{R}}
\newtheorem{thm}{Theorem}[section]

\newtheorem{corol}[thm]{Corollary}

\newtheorem{remark}[thm]{Remark}

\usepackage[inline]{enumitem}
\DeclareSIUnit{\byte}{B}
\DeclareSIUnit{\kB}{\kilo\byte}
\DeclareSIUnit{\MB}{\mega\byte}
\DeclareSIUnit{\pJ}{\pico\joule}
\DeclareSIUnit{\pJperByte}{\pJ\per\byte}

\newcommand{\CommentRight}[1]{%
  \hfill%
  \parbox[t]{0.35\linewidth}{%
      \raggedleft%
      \footnotesize%
      $\triangleright$~\begin{minipage}[t]{0.95\linewidth}%
         #1%
      \end{minipage}%
  }%
}

\title{From Silicon to Spikes: System-Wide Efficiency Gains via Exact Event-Driven Training in Neuromorphic Computing}

\author{
  Arman Ferdowsi and Atakan Aral\\
  Faculty of Computer Science, University of Vienna, Austria\\
  \texttt{arman.ferdowsi@univie.ac.at}, \texttt{atakan.aral@univie.ac.at}
}

\date{} 

\begin{document}
\maketitle

\begin{abstract}
Spiking neural networks (SNNs) promise orders-of-magnitude efficiency gains by communicating with sparse, event-driven spikes rather than dense numerical activations. However, most training pipelines either rely on surrogate-gradient approximations or require dense time-step simulations, both of which conflict with the memory, bandwidth, and scheduling constraints of neuromorphic hardware and blur precise spike timing. We introduce an analytical, event-driven learning framework that computes exact gradients for synaptic weights, programmable transmission delays, and adaptive firing thresholds, three orthogonal temporal controls that jointly shape SNN accuracy and robustness. By propagating error signals only at spike events and integrating subthreshold dynamics in closed form, the method eliminates the need to store membrane-potential traces and reduces on-chip memory traffic by up to $24\times$ in our experiments. Across multiple sequential event-stream benchmarks, the framework improves accuracy by up to $7\,\%$ over a strong surrogate-gradient baseline, while sharpening spike-timing precision and enhancing resilience to injected hardware noise. These findings indicate that aligning neuron dynamics and training dynamics with event-sparse execution can simultaneously improve functional performance and resource efficiency in neuromorphic systems.
\end{abstract}



\section{Introduction}
The rapid scaling of Artificial Intelligence (AI) workloads is colliding with hard physical limits. State-of-the-art language and vision models already draw \emph{megawatts} in large data centers and drain batteries in mobile and embedded platforms; moving activations back and forth between memory and compute has become comparable in cost to the arithmetic itself. As power, cooling, and carbon budgets shift from secondary concerns to primary design constraints, there is growing consensus that incremental efficiency gains from conventional von Neumann architectures will not be sufficient.

\emph{Neuromorphic computing} offers a qualitatively different path. By co-locating memory and compute and representing information with sparse, binary \emph{spikes}, prototype chips such as IBM's \texttt{TrueNorth} and Intel's \texttt{Loihi}/\texttt{Loihi-2} have already demonstrated compelling energy-per-inference gains on targeted applications compared to conventional GPUs and CPUs \cite{Akopyan2015TrueNorth,Davies2018Loihi,Intel2024Loihi2Brief}. These platforms execute asynchronous networks of \emph{leaky integrate-and-fire} neurons that communicate exclusively through spike events, and they expose synaptic weights, axonal delays, and neuron parameters as programmable knobs. From a systems perspective, this makes neuromorphic processors more than a biological curiosity: they are credible candidates for post-Moore efficiency and algorithm-hardware co-design.

\begin{figure}[t]
  \centering
    \includegraphics[width=0.5\linewidth]{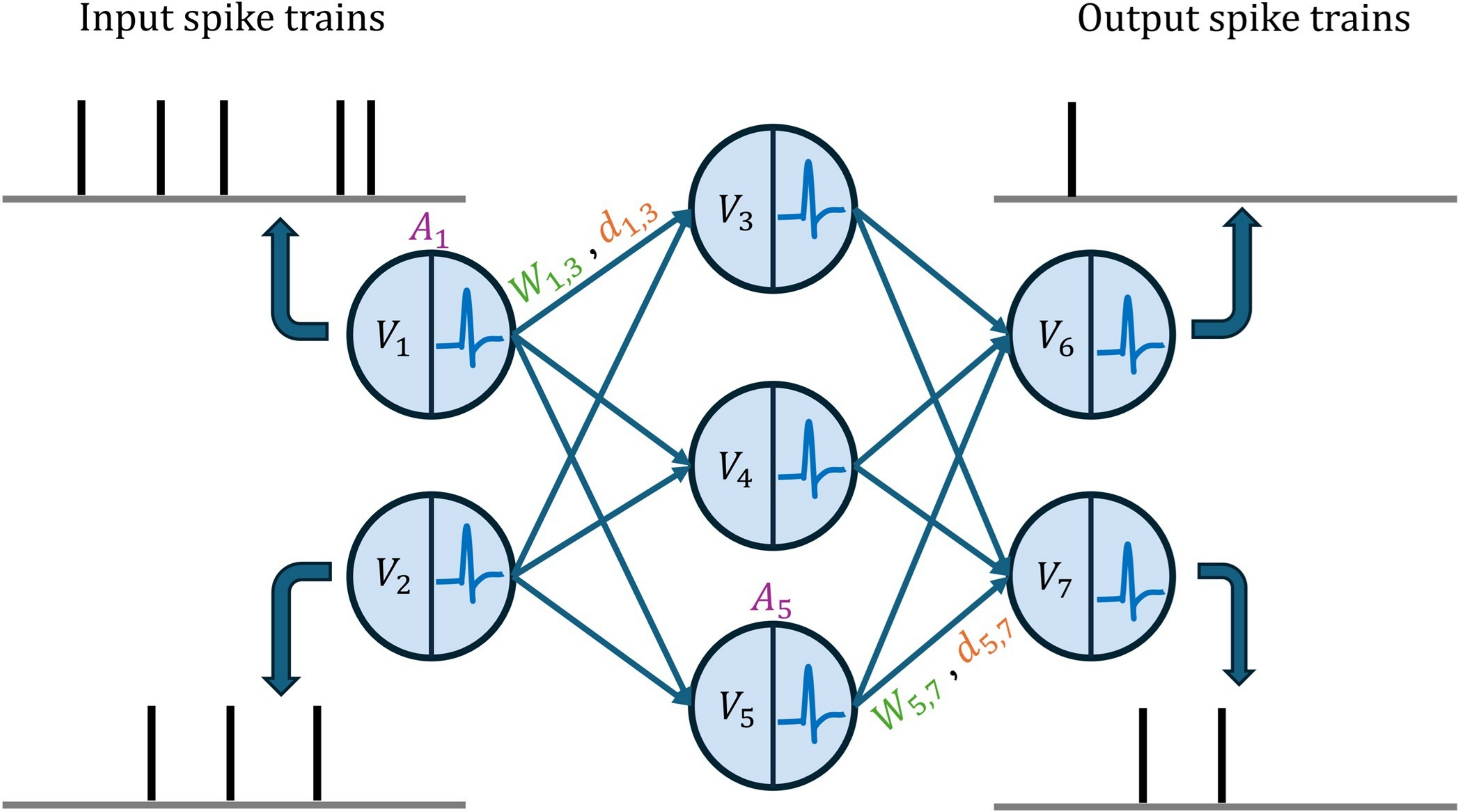}%
  \caption{{\small A schematic of a spiking neural network. $A_{i}$, $W_{ij}$, and $d_{ij}$ are trainable \emph{adaptive threshold}, \emph{weight}, and \emph{delay} parameters.}}\label{Fig:SNN}
\end{figure}

Spiking Neural Networks (SNNs), often described as the ``third generation'' of neural models \cite{Maass1997}, are the natural algorithmic counterpart to neuromorphic hardware, where information is encoded not in dense, real-valued activations but in the timing and pattern of spikes, which can be transmitted and processed in an event-driven fashion. As illustrated in \cref{Fig:SNN}, each neuron collects incoming spike trains, integrates their contributions through synaptic weights $W_{ij}$ and delays $d_{ij}$, and emits a spike once its membrane potential crosses a (possibly adaptive) firing threshold $A_j$. The resulting computations are inherently sparse in time and space: synapses that do not see spikes do not consume energy, and idle neurons do not update their state even though their membrane potential continues to decay over time \cite{Akopyan2015TrueNorth}. Under favorable conditions, this sparsity translates into orders-of-magnitude energy savings relative to dense Artificial Neural Networks (ANNs) \cite{Hammouamri2023}.

Yet despite this promising picture, there is a sharp mismatch between how SNNs \emph{run} on neuromorphic hardware and how they are typically \emph{trained} in software. Regarding the latter, the dominant training pipelines today rely on \emph{surrogate gradients} (SG): the non-differentiable spike function is replaced with a smooth pseudo-derivative, and the continuous membrane dynamics are discretized in time so that backpropagation through time (BPTT) can be applied \cite{Neftci2019,neftci2019surrogate,li2021differentiable,lian2023learnable,bittar2022surrogate}. While highly effective on many benchmarks, and scalable to deep, residual, and attention-style SNN architectures \cite{fang2021deep,yao2023attention,zhou2022spikformer}, this approach has three structural drawbacks that become increasingly problematic at system scale:
\begin{enumerate*}[label=(\roman*)]
\item dense time discretization blurs precise spike timing and entangles temporal credit assignment with arbitrary time-step choices \cite{Wunderlich2021,zhu2022training,moro2024role},
\item training requires storing and revisiting full membrane-potential traces at all timesteps, inflating memory and on-chip bandwidth \cite{hu2024high},
\item the resulting learning dynamics are misaligned with the sparse, event-driven execution model of neuromorphic processors, which never ``see'' those dense traces in hardware \cite{zhu2022training,rast2024efficient}.
\end{enumerate*}

In practice, this induces a persistent trade-off. Methods that push accuracy tend to lean on dense-time simulations and rich surrogate machinery, while methods that embrace strict event sparsity often sacrifice gradient fidelity or fall back to local, heuristic rules that are difficult to scale and to integrate into modern toolchains. Recent surveys underscore this tension: Rast et al. \cite{rast2024efficient} systematically map out the accuracy-efficiency frontier for SNN learning and identify the lack of hardware-aligned, gradient-based training as a central obstacle to broader adoption. Moser and Lunglmayr \cite{moser2024spiking}, on the other hand, emphasize that continuous-time, event-driven formulations are mathematically natural for spike-based systems and provide rigorous error bounds, but note that most practical training pipelines still ignore this structure.

A central observation behind this work is that the temporal degrees of freedom in SNNs are richer than synaptic weights alone. In addition to $W_{ij}$ scaling post-synaptic potentials, many neuromorphic platforms expose \emph{axonal transmission delays} $d_{ij}$ and \emph{adaptive firing thresholds} $A_j$ as first-class, programmable parameters. Biologically, delays govern when spikes arrive and which inputs coincide in time, while adaptive thresholds modulate a neuron's excitability based on its recent firing history, effectively implementing a form of activity-dependent memory \cite{marom2023biophysical,benda2008spike}. Theoretically, jointly controlling weights, delays, and thresholds allows SNNs to represent temporal computations that static, rate-based models cannot realize \cite{Maass1999,maass1999computing}. System-wise, all three parameter families are already reflected, in one form or another, in contemporary neuromorphic chips.

Existing learning methods, however, either ignore these temporal controls or treat them separately and approximately. Most surrogate-gradient frameworks optimize only the weights while keeping delays fixed or hand-tuned. A smaller subset adds trainable delays but still relies on discrete-time surrogates and dense simulation \cite{shrestha2018slayer,Sun2023,Hammouamri2023}. Intrinsic plasticity mechanisms such as adaptive thresholds are typically learned in isolation or via specialized architectures (e.g., LSNNs, GLIF variants, adaptive currents), again within surrogate-gradient or heuristic frameworks \cite{bellec2018long,fang2021deep,yao2022glif,bittar2024exploring}. In this light, more recent \emph{exact} event-driven approaches, such as EventProp and DelGrad, move beyond surrogates by analytically differentiating spike times in continuous time, but they cover only subsets of the parameter space (weights in EventProp \cite{Wunderlich2021} and weights plus delays in DelGrad \cite{Goeltz2023}). To the best of our knowledge, there is still no unified framework that
\begin{enumerate*}[label=(\alph*)]
\item computes \emph{exact} gradients in continuous time,
\item operates strictly at spike events, and
\item co-optimizes synaptic weights, programmable delays, and adaptive thresholds under a single, consistent calculus.
\end{enumerate*}

In this work, we address this gap by developing an \emph{analytical, event-driven learning framework} that operates directly in continuous time and provides exact gradients for these three orthogonal temporal controls. Conceptually, we treat each spike time as an implicit function of all neuro-synaptic parameters and apply an event-based form of the Implicit Function Theorem to differentiate through threshold crossings. As long as spike counts and orderings remain stable under small parameter perturbations, a mild condition that we formalize, spike times are differentiable, and their gradients can be backpropagated through the network without any surrogate smoothing or dense time discretization. Algorithmically, this yields an event-driven backpropagation scheme in which gradients are computed only at spike events, subthreshold dynamics are handled analytically, and the only state that must be stored during training is a set of spike times and synapse indices. Consequently, both computation and memory traffic scale with the number of spikes rather than wall-clock time, aligning the learning dynamics with the event-driven execution model of neuromorphic hardware.

In summary, the main contributions of this work are as follows:
\begin{enumerate}
\item \textbf{Exact event-driven gradients in continuous time.} We derive closed-form gradients for synaptic weights, axonal delays, and adaptive thresholds under a current-based LIF model, proving differentiability under simple up-crossings and giving explicit update formulas that fire \emph{only at spike events}.
\item \textbf{Algorithm and efficiency.} We formulate an event-driven backpropagation algorithm that stores only spike times (no membrane traces), so both compute and memory scale with the number of events rather than wall time.
\item \textbf{Empirical validation.} Across five event-stream benchmarks and multiple silicon back-ends, the method improves accuracy and timing precision over a strong surrogate-gradient baseline while reducing on-chip traffic and energy, demonstrating alignment between the learning rule and event-driven execution.
\end{enumerate}

We organize the paper as follows. \cref{sec:RW} briefly surveys related work. In \cref{Sec:framework}, we introduce our theoretical framework for computing exact gradients and formulate our event-driven training algorithm. In \cref{sec:exp}, we elaborate on the experimental setup and report the results. \cref{sec:con} finally summarizes our contributions and outlines potential directions for future work.

\section{Related work}
\label{sec:RW}
Neuromorphic systems place communication and memory next to compute and exploit event sparsity to reduce energy and data movement. Mature prototypes such as IBM \texttt{TrueNorth} and Intel \texttt{Loihi}/\texttt{Loihi-2} have shown that spike-based processing can deliver favorable energy/inference on selected workloads while exposing rich on-chip programmability for future learning rules \cite{Akopyan2015TrueNorth,Debole2019,Davies2018,Intel2024Loihi2Brief,ref_loihi2,davies2021advancing}. Broader surveys position neuromorphic computing as a credible path toward efficiency and algorithm-hardware co-design beyond von Neumann scaling \cite{ref_neuromorphic_survey}. From an Edge AI perspective, recent work argues that ultra-low-latency, on-device inference and learning will increasingly demand architectures that minimize memory traffic, support event-sparse computation, and integrate non-von Neumann primitives into the computing continuum \cite{kimovski2023beyond, meuser2024revisiting}. In parallel, the SNN community has clarified both algorithmic opportunities and open challenges: recent overviews emphasize efficiency/learnability trade-offs \cite{rast2024efficient}, whereas functional analyses in continuous time sharpen our understanding of spike-train representations and error bounds \cite{moser2024spiking}.

The remainder of this section provides a more focused overview of current advances in training SNNs.

\subsection{Surrogate gradients}
The dominant recipe for supervised SNN training replaces the spiking discontinuity with a smooth pseudo-derivative and then applies backpropagation through time to a discretized version of the membrane dynamics \cite{Neftci2019,neftci2019surrogate,li2021differentiable,lian2023learnable,chen2023approximate,bittar2022surrogate}. Surrogate-trained SNNs now scale to deep residual and attention-style models and even transformer-like blocks for sequential tasks \cite{fang2021deep,yao2023attention,zhou2022spikformer}, and recent theory relates surrogate estimators to stochastic automatic differentiation, clarifying when and why they work \cite{gygax2025elucidating, Buesing2011}. At the same time, the method's practical cost comes from dense time stepping and the need to materialize membrane traces, which inflates training memory and bandwidth \cite{hu2024high}. Moreover, when we care about precise spike timing, the discretization and the proxy derivative can blur temporal credit assignment \cite{zhu2022training,moro2024role}. Several studies therefore seek lighter-weight spike-timing backpropagation or alternative objectives that retain some of the timing semantics while cutting cost \cite{gong2024lightweight,yang2024maximum}. These advances motivate, but do not yet fully provide, a formulation in which gradients are computed \emph{exactly} in continuous time and emitted \emph{only at events}.

\noindent
\emph{This gap motivates learning rules that compute gradients exactly in continuous time and operate natively on spike events.}

\subsection{Temporal coding via trainable delays}
One key source of temporal expressivity in spiking systems is axonal conduction and synaptic time courses, which align spikes in time: the arrival lag of inputs shapes coincidence and integration \cite{purves2019neurosciences,dan2004spike}. Classical theory shows that even single-layer spike-response models with programmable delays can represent rich temporal functions unreachable by rate-only perceptrons, and that learning with temporal codes can be computationally powerful \cite{Maass1999,maass1999computing}. In neuromorphic sensing, delay structure is integral to tasks like motion and gesture processing \cite{orchard2015converting}. A line of work therefore targets \emph{delay learning}: local or supervised rules (e.g., STDP-inspired and ReSuMe-style) adjust delivery times to meet desired spike alignments \cite{senn2022spike,taherkhani2015dl}. Surrogate-gradient frameworks have incorporated delays as differentiable parameters (e.g., SLAYER and variants with delay caps) \cite{shrestha2018slayer,Sun2023}, while others propose structured parameterizations (dilated convolutions with learnable spacing) \cite{Hammouamri2023} or domain-specific mechanisms (optical SNNs, bioinspired motion detectors) \cite{han2021delay,grimaldi2023learning}. Strikingly, recent studies reported competitive performance with \emph{delay-only} training for certain networks \cite{grappolini2023beyond}, underscoring the computational leverage of timing.

\noindent \emph{Yet most of these methods still rely on densely sampled time grids or smoothing and thus inherit the memory and scheduling mismatch to event-driven hardware.}

\subsection{Intrinsic plasticity and adaptive thresholds}
Complementary to synaptic and axonal mechanisms, neurons adapt their excitability across spikes and timescales via threshold dynamics and after-currents. This intrinsic plasticity shapes temporal computation and robustness \cite{marom2023biophysical}. Spike-frequency adaptation and related mechanisms improve sensitivity to temporal structure and guard against runaway firing \cite{benda2008spike}. In SNN modeling, LSNNs (adaptive LIF) introduced per-spike threshold increments that decay slowly, enabling long-range credit assignment with a small, neurally plausible state \cite{bellec2018long}.
Follow-up work has learned neuron time constants or reset values to calibrate dynamics \cite{fang2021deep,perez2021neural}, incorporated gating mechanisms (GLIF) to modulate integration \cite{yao2022glif}, or preferred adaptation currents (AdLIF) over threshold modulation on certain benchmarks \cite{bittar2024exploring}. These works demonstrate that \emph{intrinsic} parameters meaningfully complement synaptic plasticity.

\noindent \emph{Nevertheless, joint optimization of delays and neuronal adaptation has received much less attention, especially in frameworks that avoid surrogates.}

\subsection{Exact event-based differentiation} 
Building on the above, a natural complementary line of work is to seek methods that compute gradients exactly in continuous time by differentiating through event discontinuities. EventProp showed that, under simple up-crossings, one can backpropagate \emph{exact} timing gradients without surrogate smoothing \cite{Wunderlich2021}. DelGrad extended this calculus to trainable synaptic delays, deriving closed-form partials for both weights and delays and demonstrating accuracy/robustness improvements together with event-driven sparsity \cite{Goeltz2023}. These works frame spike times as implicit functions of parameters and apply the implicit function theorem to obtain gradients localized at events, thereby avoiding the need to store membrane trace tensors. Our framework builds directly on this perspective and, to the best of our knowledge, is the first to provide exact event-based gradients that \emph{jointly} cover weights, delays, and \emph{adaptive thresholds} in a single training loop, thereby unifying explicit spike alignment with neuron-intrinsic excitability control. In parallel, there is growing interest in closing the algorithm/hardware loop: exact backprop variants have been prototyped on neuromorphic substrates \cite{Renner2024backpropagation}, and common intermediate representations aim to make learning rules portable across devices \cite{pedersen2024neuromorphic}.

\subsection{System and toolchain perspective}
Learning rules should reflect not only task metrics but also the realities of routing, placement, and memory on neuromorphic platforms. Compiler pipelines map SNN graphs to cores and interconnects \cite{Fang2020CompileSNN,ji2018bridge}, and event-driven simulation/measurement stacks support hardware-in-the-loop iteration \cite{lee2021neuroengine}. Large-scale studies examine how to place millions to billions of neurons while reducing NoC congestion and energy \cite{jin2023mapping,Zhu2023VLSISNN}. At the device level, the energy of moving a byte often dominates arithmetic \cite{horowitz20141}; avoiding dense state tensors is therefore crucial. Recent surveys across FPGA and mixed-signal designs echo the need for learning methods that maintain event sparsity and are portable across diverse substrates \cite{bouvier2019spiking,furber2016large,karamimanesh2025spiking,ref_neuromorphic_survey}.  \emph{Our choice to compute gradients only at spikes removes the need to store or shuttle membrane traces, aligning algorithmic cost with the events that hardware naturally processes.} Practical reports on \texttt{Loihi}/\texttt{Loihi-2} likewise highlight the importance of traffic locality and on-chip memory budgets for end-to-end efficiency \cite{davies2021advancing,ref_loihi2}.

Beyond single-device execution, recent work on distributed neuromorphic edge systems has further highlighted how event-driven computation reduces communication and energy overheads in real deployments, underscoring the need for training rules that respect hardware-level constraints and sparsity patterns \cite{ferdowsi2025distributed}.

\subsection{Synthesis and positioning}	
To summarize, bringing these threads together, prior work falls into three broad families: (i) surrogate-gradient methods that are versatile but dense in time and approximate around the discontinuity \cite{Neftci2019,lian2023learnable,bittar2022surrogate,fang2021deep,yao2023attention,gygax2025elucidating,hu2024high,moro2024role}, (ii) timing-aware models that learn \emph{delays} or \emph{adaptation}, often separately and often through surrogates or local rules \cite{Sun2023,Hammouamri2023,
grimaldi2023learning,grappolini2023beyond,marom2023biophysical,bittar2024exploring,yang2024maximum,deckers2024co}, and (iii) exact, event-driven differentiation frameworks that, so far, cover weights (EventProp) and weights+delays (DelGrad) but not neuron-intrinsic adaptation \cite{Wunderlich2021,Goeltz2023}. Our contribution extends (iii) by deriving and implementing \emph{closed-form, continuous-time} gradients for weights, delays, and adaptive thresholds in a single, event-triggered algorithm. This directly addresses the core limitations of (i) and (ii): it (a) removes dense time stepping and surrogate smoothing, (b) enables precise temporal credit assignment via spike times, and (c) matches the event-sparse execution model of contemporary neuromorphic hardware, thus lowering memory traffic and energy while improving timing precision.

Methodologically, exact event-driven gradients reconcile temporal expressivity with training efficiency: delays align spike arrivals, thresholds modulate excitability across spikes, and weights scale impulses, three orthogonal temporal controls optimized with the same analytical machinery. System-wise, event-local gradients eliminate the need to expose dense membrane traces to software, allowing SRAM-resident training buffers and reducing on-chip traffic, which is typically the dominant energy and thermal driver \cite{horowitz20141,davies2021advancing}. Complementary infrastructure trends, including portable IRs, on-device learning prototypes, and scalable placement and mapping, suggest that such learning rules can be incorporated into real toolchains \cite{pedersen2024neuromorphic,Renner2024backpropagation,jin2023mapping,Zhu2023VLSISNN}.

Viewed together, these strands of work expose a three-way tension between respecting the true spiking dynamics, exploiting rich temporal codes through delays and neuronal adaptation, and staying aligned with event-sparse neuromorphic hardware. Existing methods typically optimize at most two of these aspects at once. The framework we introduce next is deliberately shaped around this triangle, taking spike events themselves as the basic unit of both computation and learning and, in doing so, opens up a new operating point in the design space of SNN training rules.

\section{Theoretical framework (Training Procedure)}
\label{Sec:framework}

\subsection{Network topology and modeling}
\label{Section:model}
We focus on a current-based \emph{Leaky Integrate-and-Fire} (LIF) neuron model, aligned with the default on-chip neuron abstraction on contemporary neuromorphic processors. We consider a layered (feedforward) SNN with $L$ layers, neuron set
$\mathcal{V}=\bigcup_{\ell=1}^{L}\mathcal{V}_\ell$
and directed synapses
\(\mathcal{E}\subseteq\mathcal{V}\times\mathcal{V}\).
For each synapse $(i\!\to\!j)\in\mathcal{E}$ we denote its synaptic weight by
$w_{ij}\in\mathbb{R}$ and its (non-negative) transmission delay by
$d_{ij}\in\mathbb{R}_{\ge 0}$, meaning that if neuron $i$ emits its $f$-th spike at time $t_i^{(f)}$, then neuron $j$ receives that spike at time $t_i^{(f)} + d_{ij}$. The pre-synaptic spike train of neuron $i$ is modeled as
\[
S_i(t) \;=\; \sum_{f} \delta\!\bigl(t - t_i^{(f)}\bigr),
\]
where $\delta(\cdot)$ is the Dirac delta distribution and we assume a finite number of spikes in any finite time window.

Then, each post-synaptic membrane potential $V_j(t)$ evolves, between reset events, according to the current-based LIF dynamics
\begin{equation}
  \tau_m \dot V_j(t)
  \;=\;
  -\bigl(V_j(t)-V_{\text{rest}}\bigr)
  + \sum_{i,f:\,(i\to j)\in\mathcal{E}} w_{ij}\,
      \delta\!\bigl(t - [t_{i}^{(f)} + d_{ij}]\bigr),
  \label{eq:lif}
\end{equation}
where $\tau_m>0$ is the membrane time constant, $V_{\text{rest}}$ is the resting potential (often taken as $0$ for simplicity), and $t_i^{(f)}$ denotes the $f$‑th spike time of neuron $i$. 

\cref{eq:lif} is a first-order linear ordinary differential equation in $V_j$ driven by a sum of impulse inputs. It is convenient to subtract the resting potential and write
\(y_j(t)=V_j(t)-V_{\text{rest}}\). Then $y_j$ satisfies the linear time-invariant (LTI) relation
\[
  \tau_m \dot y_j(t) + y_j(t)
  \;=\;
  \sum_{i,f:\,(i\to j)\in\mathcal{E}} w_{ij}\,
      \delta\!\bigl(t - [t_{i}^{(f)} + d_{ij}]\bigr).
\]
A standard way to solve such an LTI system with impulsive input is via its \emph{Green's function} (impulse response) \cite{duffy2015green}. For the homogeneous equation
\(\tau_m \dot y+y=0\) the solution decays exponentially. The unique causal impulse response $\omega(\tau)$ solving
\[
  \tau_m \dot \omega(\tau) + \omega(\tau) = \delta(\tau),\qquad
  \omega(\tau)=0\ \text{for }\tau<0
\]
is
\begin{align}
\omega(\tau)\;=\;
\begin{cases}
\dfrac{1}{\tau_m}\,e^{-\tau/\tau_m}, & \tau\ge 0,\\[4pt]
0, & \tau<0,
\end{cases}
\label{eq_omega}
\end{align}
which is the standard causal post-synaptic potential (PSP) kernel for an instantaneous synaptic input in a simple LIF model \cite{gerstner2002spiking,perestyuk1995impulsive}.

Because the system is linear and time-invariant, the response to a sum of impulses is the sum of time-shifted, scaled copies of $\omega$. Assuming that at the beginning of each inter-spike interval the membrane is at rest, $V_j(t_0)=V_{\text{rest}}$, the subthreshold solution on that interval can be written as
\begin{align}
V_j(t)
\;=\;
V_{\text{rest}}
\;+\;
\sum_{i,f:\,(i\to j)\in\mathcal{E}}
  w_{ij}\,\omega\!\bigl(t - [t_i^{(f)} + d_{ij}]\bigr), 
\label{eq:V_as_exp}
\end{align}
where the sum runs over all presynaptic spikes $(i,f)$ whose arrival times
$t_i^{(f)}+d_{ij}$ lie in the past of $t$. In \cref{eq:V_as_exp}, the delay $d_{ij}$ appears only as a time shift in the kernel argument, while the weight $w_{ij}$ scales the kernel amplitude. In the full LIF model, the same expression is used \emph{piecewise} between reset events: at a spike of neuron $j$, the membrane is reset (or clamped) according to the reset rule, and the subsequent subthreshold evolution again follows \cref{eq:lif} with the new initial condition. All derivatives used for learning are evaluated on this \emph{pre-reset} trajectory at the moment of threshold crossing.

For instance, intuitively, changing a synaptic delay $d_{ij}$ shifts the arrival time of a presynaptic spike and thus can advance or delay the moment when $V_j(t)$ first reaches threshold. If two key inputs arrive more synchronously, neuron $j$ may reach threshold earlier (a smaller $t_j$) or with fewer total spikes; if their arrivals are sufficiently desynchronized, $j$ may not fire at all. \cref{fig:effect} illustrates this phenomenon and the way it leads to a well-defined timing gradient via implicit differentiation.

\begin{figure}[t]
\centering
  \begin{tikzpicture}[>=stealth,thick,scale=1.1, transform shape, font=\footnotesize]
    \draw[->] (0,0) -- (6.3,0) node[below] {\textbf{time}};
    \draw[->] (0,0) -- (0,3.0) node[left] {$V_j(t)$};

    \draw[dashed] (0,2) -- (6,2) node[pos=1,right] {threshold};

    \draw[|-|,gray!60!black] (1,-0.2) -- (1,0.2);
    \node[below] at (1,-0.2) {$t_i$};

    \draw[|-|,gray!60!black] (2,-0.2) -- (2,0.2);
    \node[below] at (2,-0.2) {$t_i + d_{ij}$};

    \draw[blue,samples=100,domain=2.88:5.5,smooth,variable=\x]
      plot (\x,{1.0 + 1.2*(1 - exp(-(1.5*(\x-2.95))))});
    \draw[|-|,blue] (4.2,-0.2) -- (4.2,0.2);
    \node[below,blue] at (4.2,-0.2) {$t_j$};

    \draw[red,dashed,samples=100,domain=2.6:4.3,smooth,variable=\x]
      plot (\x,{1.0 + 1.5*(1 - exp(-(1.5*(\x-2.6))))});
    \draw[|-|,red] (3.3,-0.2) -- (3.3,0.2);
    \node[below,red] at (3.3,-0.2) {$t_j - \Delta t_j$};

    \draw[->,red] (3.4,1.3) .. controls (3,1.3) and (2.8,1.3) .. (2.8,1.3)
      node[midway,xshift=-40pt,teal]
      {\(\downarrow d_{ij}\;\Rightarrow\;\downarrow t_j\)};

    \draw[->] (1,0.7) -- (1,0.1)
      node[midway,right=1pt,gray]{\scriptsize arrival of presyn. spike};
    
    \node[align=center, font=\tiny] at (5.1,0.9)
    {
     \(\displaystyle
      \frac{\partial t_j}{\partial d_{ij}} \;=\;
      -\,\frac{\partial_{d_{ij}} V_j(t_j)}{\partial_t V_j(t_j)}
     \)
    };
  \end{tikzpicture}
  \caption{\small
  Illustration of how a synaptic delay $d_{ij}$ affects the spike time $t_j$.
  The neuron fires when $V_j(t)$ reaches the dashed threshold line. Decreasing
  $d_{ij}$ advances the arrival of the presynaptic input, shifting the
  threshold crossing earlier from $t_j$ to $t_j-\Delta t_j$. At the original
  time $t_j$, the voltage may actually be \emph{lower} due to additional
  decay; it is the \emph{shift in crossing time}, not a larger
  $V_j(t_j)$, that causes the earlier spike. The schematic derivative shown
  is the usual implicit-function formula for a static threshold; with a
  dynamic (adaptive) threshold $\nu_j(t)$, the denominator becomes the local
  crossing slope \(\partial_t[V_j(t)-\nu_j(t)]|_{t=t_j^-}\), see
  \cref{eq:threshold_condition2}.
  }
  \label{fig:effect}
\end{figure}

\paragraph{Static threshold and spike times.}
For the moment, suppose neuron $j$ fires whenever its potential crosses a \emph{constant} threshold $V_{\text{thr}}$ from below. If $j$ fires exactly once in a time window and we denote that spike time by $t_j$, then the crossing condition can be written as
\begin{align}
\label{eq:threshold_condition}
V_j(t_j) \;=\; V_{\text{thr}}
\quad\text{and}\quad
\partial_t\!\bigl(V_j(t)-V_{\text{thr}}\bigr)\big|_{t=t_j^-} \;>\; 0.
\end{align}
The derivative is taken from the left, because at $t_j$ the membrane potential is immediately reset and the trajectory becomes discontinuous. The sign constraint enforces an \emph{up-crossing}; otherwise spike creation/annihilation events may occur under infinitesimal parameter perturbations, which is precisely what we must exclude to obtain differentiable spike times later (cf.\ \cref{Sec:IDST}).

The spike-generation nonlinearity is a Heaviside step function: a neuron is quiescent below threshold and emits a spike as soon as the membrane crosses the threshold. If one differentiates this step nonlinearity directly, the derivative is zero almost everywhere and undefined at the crossing, which obstructs straightforward gradient-based learning \cite{auge2021survey}. Surrogate-gradient methods bypass this by replacing the exact discontinuity with a smooth pseudo-derivative \cite{Neftci2019,neftci2019surrogate}, at the cost of dense time discretization and an approximate treatment of spike timing. In contrast, our approach exploits the fact that each spike time $t_j$ can be viewed as an \emph{implicit} function of all parameters that influence $V_j$: the crossing condition in \cref{eq:threshold_condition} defines $t_j$ implicitly through the equation
\(V_j(t_j)=V_{\text{thr}}\). Under mild regularity conditions, notably that the number and order of spikes do not change under small parameter perturbations and that the crossing is simple, the Implicit Function Theorem guarantees that $t_j$ is locally differentiable with respect to all such parameters (see \cref{Sec:IDST} for a formal statement).

\paragraph{Adaptive threshold dynamics.}
Beyond fixed thresholds, many biological neurons exhibit spike-frequency adaptation: after firing, they become temporarily less excitable before gradually returning to baseline. A common phenomenological model captures this via an \emph{adaptation variable} $a_j(t)$ that is incremented after each spike and decays exponentially in time. We adopt this structure and treat the adaptation strength as a trainable parameter.

Specifically, each neuron $j$ is equipped with a dynamic firing threshold
\(\nu_j(t)\) of the form
\[
  \nu_j(t) \;=\; \nu_0 + a_j(t),
\]
where $\nu_0$ is a fixed baseline threshold and $a_j(t)$ is governed by a hybrid system:
\[
  \tau_a \dot a_j(t) = -a_j(t)
\]
between spikes, with jumps of size $A_j$ at each spike of neuron $j$. When neuron $j$ fires its $f$-th spike at time $t_j^{(f)}$, we impose
\[
     a_j\bigl(t_j^{(f)+}\bigr) 
     \;=\; a_j\bigl(t_j^{(f)-}\bigr) 
     \;+\; A_j,
\]
where $t_j^{(f)\pm}$ denote the instants just before and just after the spike, and $A_j$ is a \emph{trainable} adaptation amplitude controlling how strongly the neuron's threshold is raised per spike. At the beginning of a trial we set $a_j(0)=0$, so before the very first spike we have $a_j(t_j^{(1)-})=0$ and hence $a_j(t_j^{(1)+})=A_j$. Solving the linear ODE with these jump conditions yields, for any time $t$ after the first spike of neuron $j$,
\begin{align}
a_j(t)\;=\;A_j\sum_{g:\,t_j^{(g)}<t}\exp\!\Bigl(-\tfrac{t-t_j^{(g)}}{\tau_a}\Bigr),
\qquad
\nu_j(t)\;=\;\nu_0+a_j(t). 
\end{align}
Thus, the instantaneous threshold is a weighted sum of exponentially decaying contributions from the neuron's own past spikes. When $A_j=0$ or before the first spike, we have $\nu_j(t)\equiv\nu_0$, recovering the static-threshold case.

With this adaptive threshold, the spike time $t_j$ of neuron $j$ is defined by the dynamic crossing condition
\begin{align}
\label{eq:threshold_condition2}
V_j(t_j) \;=\; \nu_j(t_j)
\quad\text{and}\quad
\partial_t\!\bigl(V_j(t)-\nu_j(t)\bigr)\Big|_{t=t_j^-} \;>\; 0. 
\end{align}
Again, the derivative is taken on the pre-reset trajectory. Note that \cref{eq:threshold_condition2} reduces to \cref{eq:threshold_condition} when $\nu_j(t)$ is constant in time (e.g., $\nu_j(t)\equiv V_{\text{thr}}$).

\paragraph{Parameterization for learning.}
Collecting all trainable parameters in the set
\[
\Theta \;=\; \bigl\{W=\{w_{ij}\},\; D=\{d_{ij}\},\; A=\{A_j\}\bigr\},
\]
we can make the dependence of the membrane potential on $\Theta$ explicit by writing
\begin{align}
V_j\bigl(t,\Theta\bigr)
\;=\;
V_{\text{rest}}
\;+\;
\sum_{i,f:\,(i\to j)\in\mathcal{E}}
  w_{ij}\,\omega\!\bigl(t - [t_i^{(f)} + d_{ij}]\bigr).
\label{eq:V_as_exp2}
\end{align}
Spike times $t_j^{(f)}(\Theta)$ are then defined implicitly as the solutions of
\begin{align}
V_j\bigl(t_j^{(f)}(\Theta),\Theta\bigr)=\nu_j\bigl(t_j^{(f)}(\Theta)\bigr)
\label{eq:ThreExt}
\end{align}
together with the up-crossing condition in \cref{eq:threshold_condition2}. The central goal of this work is to compute \emph{exact} gradients of a task loss with respect to $W$, $D$, and $A$ by differentiating these implicitly defined spike times in continuous time. The next subsection develops the corresponding event-driven calculus.

\subsection{Implicit differentiation of spike times}
\label{Sec:IDST}
When training an SNN, one typically defines a \emph{global loss function} $L$ reflecting how well the network performs on some task. For instance, if the network must classify inputs, $L$ might measure cross-entropy between the desired label and the output derived from spike timing in the output layer. Or if the network must match a desired spike train, $L$ might be the sum of squared differences between actual spike times and target spike times. Regardless, in a feedforward SNN, the only ``outputs'' are spike times (and possibly firing rates, even though firing rates, in turn, can be functions of spike times). Consequently, one can view the entire network's behavior as determined by the \emph{set of spike times} $\{t_k \}$. Formally, we can write:
\[
L = L\bigl(\{t_k\}\bigr) \quad \text{with} \quad t_k=f(\Theta),
\]
where $\{t_k\}$ are spike times throughout the network, including $t_j$.

In order to achieve a training procedure, one needs to be able to ultimately compute the derivative of the Loss function with respect to all parameters across all layers in a backpropagation manner. For this, applying the chain rule with multiple spikes of neuron \(j\) yields, for each \(\theta_{ij}\in\Theta\),
\[
\frac{\partial L }{\partial \theta_{ij}}
\;=\;
\sum_{f}\,\frac{\partial L}{\partial  t_j^{(f)}} \cdot \frac{\partial t_j^{(f)}}{\partial \theta_{ij}}\,,
\]
where the sum runs over all spikes of neuron \(j\) in the window \([0,T]\). Assuming that $L$ is itself differentiable in $t_j$, it remains to compute the derivative of the spike time $t_j$ in terms of the parameters, which is the primary challenge due to the fact that such a derivative does not globally exist. Note, however, that from the threshold condition \cref{eq:ThreExt}, we have

\begin{equation}
\label{eq:threexp}
V_j\bigl(t_j(\Theta),\Theta\bigr)\;-\;\nu_j\!\bigl(t_j(\Theta),\Theta\bigr) \;=\; 0.
\end{equation}

Since each $t_j(\Theta)$ is an implicit function of the parameters, we can
apply the Implicit Function Theorem \cite{krantz2002implicit} to the
\emph{pre-reset} dynamics. Concretely, we define
\[
F(t,\Theta)\;:=\;V_j(t,\Theta)\;-\;\nu_j(t,\Theta),
\]
where both $V_j$ and $\nu_j$ are understood as the left-continuous,
subthreshold trajectories evaluated just \emph{before} any spike-triggered
reset or adaptation jump (i.e., using the same ``shadow'' dynamics that
were used to define the threshold-crossing times in
\cref{eq:ThreExt}). The Implicit Function Theorem then states that
$t_j(\Theta)$ is differentiable with respect to~$\Theta$ provided that
(i) $F$ is continuous and continuously differentiable in $(t,\Theta)$ in a
region where the spike pattern (number and order of spikes) does not
change, and (ii) the crossing is \emph{simple}, in the sense that
\[
\partial_t F\bigl(t_j(\Theta),\Theta\bigr)\;\neq\;0,
\]
i.e., the pre-reset trajectory crosses the threshold with nonzero slope.
The following theorem formalizes this.

\newpage
\begin{thm}[Differentiable spike times under simple crossings]\label{thm:cond}
Fix a neuron $j$ and a compact time window $[0,T]$. Let $\Theta\subset\IR^p$ be an open parameter set and define
\[
F:(0,T)\times\Theta\to\IR,\qquad
F(t,\theta)\;=\;V_j(t,\theta)\;-\;\nu_j(t,\theta),
\]
where $\theta$ stacks all trainable parameters that influence $F$ (e.g., synaptic weights, delays, and adaptation parameters). Assume:
\begin{enumerate}
\item $F$ is $\mathcal C^1$ in a neighbourhood of each zero $(t_0^{(k)},\theta_0)$ when the time derivative $\partial_t F$ is interpreted as the (pre-reset) left derivative in $t$. In particular, if synapses are instantaneous, assume that no zero $t_0^{(k)}$ coincides with a presynaptic arrival time; with a finite synaptic rise time, $\omega\in\mathcal C^1$ and this is automatic.
\item For some $\theta_0\in\Theta$, the equation $F(t,\theta_0)=0$ has exactly $m$ distinct simple roots $t_0^{(1)}<\cdots<t_0^{(m)}$ in $(0,T)$, i.e.,
      \[
      F\bigl(t_0^{(k)},\theta_0\bigr)=0
      \quad\text{and}\quad
      \partial_t F\bigl(t_0^{(k)},\theta_0\bigr)\neq 0,\qquad k=1,\dots,m.
      \]
\item (\emph{Gap condition}) There exists $\delta>0$ such that
      \[
      |F(t,\theta_0)|\;\ge\;\delta\quad\text{for all}\quad
      t\in[0,T]\setminus\bigcup_{k=1}^m\bigl(t_0^{(k)}-\delta,\,t_0^{(k)}+\delta\bigr),
      \]
      where the union is understood to be intersected with $(0,T)$ if necessary.
\end{enumerate}
Then, there exist an open neighbourhood $U\subset\Theta$ of $\theta_0$ and unique $\mathcal C^1$ maps $t^{(k)}:U\to(0,T)$, $k=1,\dots,m$, such that
\[
F\bigl(t^{(k)}(\theta),\theta\bigr)=0,\qquad
t^{(1)}(\theta)<\cdots<t^{(m)}(\theta),\qquad \forall\,\theta\in U.
\]
In particular, the spike \emph{pattern} (count and order) is invariant throughout $U$, and for each $k$,
\[
\nabla_\theta t^{(k)}(\theta)\;=\;-\frac{\nabla_\theta F(t,\theta)}{\partial_t F(t,\theta)}\Bigg|_{t=t^{(k)}(\theta)}.
\]
\end{thm}

\begin{proof}
\emph{Step 1: Local continuation of each simple root.}
Fix $k\in\{1,\dots,m\}$. By Assumptions~1--2, $F$ is continuously differentiable in a neighbourhood of $(t_0^{(k)},\theta_0)$ and $\partial_t F(t_0^{(k)},\theta_0)\neq 0$. The Implicit Function Theorem (IFT) therefore yields open intervals $I_k\subset(0,T)$ containing $t_0^{(k)}$ and open sets $U_k\subset\Theta$ containing $\theta_0$, together with a unique $\mathcal C^1$ map
\begin{flalign}
\tau_k:U_k\to I_k \ \ & \text{such that} \nonumber \\ &
F\bigl(\tau_k(\theta),\theta\bigr)=0,\ \forall\,\theta\in U_k,\quad
\tau_k(\theta_0)=t_0^{(k)}. \nonumber &
\end{flalign}
Moreover, the IFT gives
\[
\nabla_\theta \tau_k(\theta)\;=\;-\frac{\nabla_\theta F(t,\theta)}{\partial_t F(t,\theta)}\Bigg|_{t=\tau_k(\theta)}.
\]

\emph{Step 2: Global uniqueness and pattern invariance in a common neighbourhood.}
By Assumption~3, there exists $\delta>0$ such that
\[
|F(t,\theta_0)|\;\ge\;\delta
\quad\text{for all}\quad
t\in[0,T]\setminus \bigcup_{k=1}^m \bigl(t_0^{(k)}-\delta,\,t_0^{(k)}+\delta\bigr).
\]
For each $k$, the neighbourhood $I_k$ from Step~1 is an open interval
containing $t_0^{(k)}$, so by shrinking $\delta$ if necessary we may assume
that
\[
I_k'\;:=\;\bigl(t_0^{(k)}-\tfrac{\delta}{2},\,t_0^{(k)}+\tfrac{\delta}{2}\bigr)
\subset I_k
\]
for all $k$, and that the $I_k'$ are pairwise disjoint.  Each $I_k'$ then
contains exactly one zero $t_0^{(k)}$ of $F(\cdot,\theta_0)$.

Next, use the joint continuity of $F$ in $(t,\theta)$ and the compactness
of $[0,T]$ to obtain a neighbourhood $U\subset\bigcap_k U_k$ of $\theta_0$
such that
\begin{align}
\bigl|F(t,\theta)-F(t,\theta_0)\bigr|\;<\;\tfrac{\delta}{2} \nonumber
\end{align}
whenever
\begin{align}
\theta\in U,\;
t\in[0,T]\setminus \bigcup_{k=1}^m I_k'. \nonumber
\end{align}
By Assumption~3, $|F(t,\theta_0)|\ge\delta$ on
$[0,T]\setminus\bigcup_k I_k'$, hence
$|F(t,\theta)|\ge \tfrac{\delta}{2}>0$ on the same set for all $\theta\in U$.
Therefore, for every $\theta\in U$, all zeros of $F(\cdot,\theta)$ lie
inside the disjoint union $\bigcup_k I_k'$, and each $I_k'$ contains
exactly one zero.

Define $t^{(k)}(\theta):=\tau_k(\theta)$ for $\theta\in U$.  Then each
$t^{(k)}$ is $\mathcal C^1$, satisfies
$F\bigl(t^{(k)}(\theta),\theta\bigr)=0$, and, because the $I_k'$ are
disjoint and ordered, preserves the ordering
$t^{(1)}(\theta)<\cdots<t^{(m)}(\theta)$ for all $\theta\in U$.  This
establishes spike count/order invariance and uniqueness of the
continuation.

\emph{Step 3: Gradient formula.}
The derivative identity follows directly from the local IFT in Step~1: differentiating $F\bigl(t^{(k)}(\theta),\theta\bigr)=0$ with respect to $\theta$ and solving for $\nabla_\theta t^{(k)}(\theta)$ yields
\[
\nabla_\theta t^{(k)}(\theta)\;=\;-\frac{\nabla_\theta F(t,\theta)}{\partial_t F(t,\theta)}\Bigg|_{t=t^{(k)}(\theta)},
\]
which is well-defined because $\partial_t F\bigl(t^{(k)}(\theta),\theta\bigr)\neq 0$ by continuity and Assumption~2. \qedhere
\end{proof}

\paragraph{Remarks.}
(i) The theorem is agnostic to how $\nu_j$ depends on $\theta$: as long as
the resulting $F(t,\theta)=V_j(t,\theta)-\nu_j(t,\theta)$ satisfies
Assumption~1 (in particular, is $\mathcal C^1$ in a neighbourhood of each
zero when viewed with pre-reset dynamics), the conclusions hold.  This
includes the adaptive-threshold model from \cref{Section:model}, because on
each inter-spike interval $\nu_j(t)$ depends on finitely many earlier spike
times $t_j^{(g)}(\theta)$ which are themselves $\mathcal C^1$ in $\theta$
by induction over $g$. (ii) The gap condition is mild and standard: it rules out the measure-zero case of root creation/annihilation under an infinitesimal parameter change. (iii) The statement applies to any fixed neuron $j$ and any finite spike window $[0,T]$.

Having said this, \cref{thm:cond} paves the way for an exact approach to computing the derivatives incorporated into an event-driven back propagation framework:

 \begin{thm}[Exact event-driven gradients with adaptive thresholds]
\label{Prop:1}
Let $F(t):=V_j(t)-\nu_j(t)$ and consider a spike of neuron $j$ at time $t_j^{(f)}$ ($f\ge 1$) in $(0,T)$.
Assume a simple up‑crossing of $F$ at $t_j^{(f)}$, i.e.,
\[
F\bigl(t_j^{(f)}\bigr)=0
\quad\text{and}\quad
\partial_t F\bigl(t_j^{(f)}\bigr)>0,
\]
and that $t_j^{(f)}\neq t_i^{(n)}+d_{ij}$ for all presynaptic events, so that $\omega'$ is well defined at all arguments used below.
On the current inter-spike interval, the membrane potential of neuron $j$ can be written as
\[
V_j(t)
\;=\;
V_{\text{rest}}
\;+\;
\sum_{k,n} w_{kj}\,\omega\!\bigl(t-[t_k^{(n)}+d_{kj}]\bigr),
\]
and the adaptive threshold at time $t$ as
\[
\nu_j(t)
\;=\;
\nu_0
+
A_j\sum_{g:\,t_j^{(g)}<t}\exp\!\Bigl(-\tfrac{t-t_j^{(g)}}{\tau_a}\Bigr).
\]
Then, for any scalar parameter $\theta$ that influences neuron $j$ only through $V_j$ and/or $A_j$ (i.e., locally via $w_{\cdot j}$, $d_{\cdot j}$ and $A_j$), the derivative of the $f$-th spike time $t_j^{(f)}$ with respect to $\theta$ satisfies
{\small
\begin{flalign}
\label{eq:ift-template}
&\frac{\partial t_j^{(f)}}{\partial \theta}
\;=\;
-\frac{\partial_\theta F(t,\Theta)}{\partial_t F(t,\Theta)}\Bigg|_{t=t_j^{(f)}}
\;=\; \nonumber \\
& {\tiny
\frac{
\displaystyle
-\partial_\theta V_j\bigl(t_j^{(f)},\Theta\bigr)
+ \partial_\theta A_j\sum_{g=1}^{f-1}\exp\!\Bigl(-\tfrac{t_j^{(f)}-t_j^{(g)}}{\tau_a}\Bigr)
+\dfrac{A_j}{\tau_a}\sum_{g=1}^{f-1}\exp\!\Bigl(-\tfrac{t_j^{(f)}-t_j^{(g)}}{\tau_a}\Bigr)\,
\frac{\partial t_j^{(g)}}{\partial \theta}
}{
\displaystyle
V'_j\!\bigl(t_j^{(f)-}\bigr)
+\frac{A_j}{\tau_a}
\sum_{g=1}^{f-1}\exp\!\Bigl(-\tfrac{t_j^{(f)}-t_j^{(g)}}{\tau_a}\Bigr)
},}&
\end{flalign}}
where $V'_j(t_j^{(f)-})$ denotes the left derivative of $V_j$ at $t_j^{(f)}$, and we interpret the sums over $g$ as empty (and hence equal to zero) when $f=1$.

In particular, for any incoming synapse $(i\!\to\!j)$ and any spike index $f\ge1$ we have
{\small
\begin{flalign}
\label{eq:deriv1}
&\frac{\partial t_j^{(f)}}{\partial d_{ij}} 
= \nonumber \\ &
\frac{
\displaystyle
 w_{ij}\,\sum_{n} \omega'\!\bigl(t_j^{(f)} - [t_i^{(n)} + d_{ij}]\bigr)
 + \dfrac{A_j}{\tau_a}\,\sum_{g=1}^{f-1}\exp\!\Bigl(-\tfrac{t_j^{(f)}-t_j^{(g)}}{\tau_a}\Bigr)\,
      \frac{\partial t_j^{(g)}}{\partial d_{ij}}
}{
\displaystyle
\sum_{k,n} w_{kj}\,\omega'\!\bigl(t_j^{(f)} - [t_k^{(n)} + d_{kj}]\bigr)
 \;-\; \nu'_j\bigl(t_j^{(f)}\bigr)
},&
\end{flalign}}
{\small
\begin{flalign}
\label{eq:deriv2}
&\frac{\partial t_j^{(f)}}{\partial w_{ij}} 
= \nonumber \\ &
\frac{
\displaystyle
 -\sum_{n} \omega\!\bigl(t_j^{(f)} - [t_i^{(n)} + d_{ij}]\bigr)
 + \dfrac{A_j}{\tau_a}\,\sum_{g=1}^{f-1}\exp\!\Bigl(-\tfrac{t_j^{(f)}-t_j^{(g)}}{\tau_a}\Bigr)\,
      \frac{\partial t_j^{(g)}}{\partial w_{ij}}
}{
\displaystyle
\sum_{k,n} w_{kj}\,\omega'\!\bigl(t_j^{(f)} - [t_k^{(n)} + d_{kj}]\bigr)
 \;-\; \nu'_j\bigl(t_j^{(f)}\bigr)
},
& 
\end{flalign}}
and, writing $t_j^{(f)}$ explicitly to emphasize the dependence on $A_j$,
{\small
\begin{flalign}
& \frac{\partial t_j^{(1)}}{\partial A_j} \;=\; 0, \nonumber \\[2pt] &
\frac{\partial t_j^{(f)}}{\partial A_j}
\;=\;
\frac{
\displaystyle \sum_{g=1}^{f-1}\exp\!\Bigl(-\tfrac{t_j^{(f)}-t_j^{(g)}}{\tau_a}\Bigr)
\;+\;
\displaystyle \frac{A_j}{\tau_a}\sum_{g=1}^{f-1}
\exp\!\Bigl(-\tfrac{t_j^{(f)}-t_j^{(g)}}{\tau_a}\Bigr)\,\frac{\partial t_j^{(g)}}{\partial A_j}
}{
\displaystyle V'_j\!\bigl(t_j^{(f)-}\bigr)
\;+\;\frac{A_j}{\tau_a}\,\sum_{g=1}^{f-1}\exp\!\Bigl(-\tfrac{t_j^{(f)}-t_j^{(g)}}{\tau_a}\Bigr)
},
\ \ f\ge 2. 
\label{eq:deriv3}
&
\end{flalign}
}
For $f{=}1$ the sums in \cref{eq:deriv3} are empty, hence $\partial t_j^{(1)}/\partial A_j=0$.
\end{thm}

\begin{proof}
All time derivatives are understood as left derivatives on the pre-reset trajectory,
denoted $V'_j(t^-)$.

\paragraph{General form \cref{eq:ift-template}.}
Let $\theta$ be any scalar parameter that acts locally on neuron $j$, and consider the $f$-th spike of $j$. The corresponding threshold-crossing condition can be written explicitly as
\begin{align}
&F^{(f)}\bigl(t_j^{(1)},\dots,t_j^{(f)},\theta\bigr)
= \nonumber \\
& V_j\!\bigl(t_j^{(f)},\theta\bigr)
-\nu_0
- A_j(\theta)\sum_{g=1}^{f-1}\exp\!\Bigl(-\tfrac{t_j^{(f)}-t_j^{(g)}}{\tau_a}\Bigr) =0.
\label{eq:Ff-def}
\end{align}
On the present inter-spike interval, $V_j(t,\theta)$ depends on $t$ and $\theta$ but, due to the reset-to-$V_{\text{rest}}$ assumption and stability of the spike pattern, not on earlier spike times $t_j^{(g)}$ for $g<f$.

Differentiating \cref{eq:Ff-def} with respect to~$\theta$ and applying the chain rule gives
\begin{align*}
0
&=
\frac{dF^{(f)}}{d\theta} \\
&=
\partial_{\theta}F^{(f)}
+ \partial_{t^{(f)}}F^{(f)}\,\frac{\partial t_j^{(f)}}{\partial\theta}
+ \sum_{g=1}^{f-1}\partial_{t^{(g)}}F^{(f)}\,\frac{\partial t_j^{(g)}}{\partial\theta},
\end{align*}
where
\begin{align*}
\partial_{\theta}F^{(f)}
=
\partial_{\theta}V_j\bigl(t_j^{(f)},\theta\bigr)
-
\partial_{\theta}A_j(\theta)\sum_{g=1}^{f-1}\exp\!\Bigl(-\tfrac{t_j^{(f)}-t_j^{(g)}}{\tau_a}\Bigr),\\
\partial_{t^{(f)}}F^{(f)}
=
V'_j\!\bigl(t_j^{(f)-}\bigr)
+\frac{A_j(\theta)}{\tau_a}\sum_{g=1}^{f-1}\exp\!\Bigl(-\tfrac{t_j^{(f)}-t_j^{(g)}}{\tau_a}\Bigr), \\
\partial_{t^{(g)}}F^{(f)}
=
-\frac{A_j(\theta)}{\tau_a}
\exp\!\Bigl(-\tfrac{t_j^{(f)}-t_j^{(g)}}{\tau_a}\Bigr),\qquad g<f. 
\end{align*}
Substituting these into the chain-rule expression and solving for $\partial t_j^{(f)}/\partial\theta$ yields
\begin{align*}
&\left[
V'_j\!\bigl(t_j^{(f)-}\bigr)
+\frac{A_j(\theta)}{\tau_a}\sum_{g=1}^{f-1}\exp\!\Bigl(-\tfrac{t_j^{(f)}-t_j^{(g)}}{\tau_a}\Bigr)
\right]\frac{\partial t_j^{(f)}}{\partial\theta} =
\\
&\qquad
-\partial_{\theta}V_j\bigl(t_j^{(f)},\theta\bigr)
+\partial_{\theta}A_j(\theta)\sum_{g=1}^{f-1}\exp\!\Bigl(-\tfrac{t_j^{(f)}-t_j^{(g)}}{\tau_a}\Bigr)
\\
&\qquad
+\frac{A_j(\theta)}{\tau_a}\sum_{g=1}^{f-1}\exp\!\Bigl(-\tfrac{t_j^{(f)}-t_j^{(g)}}{\tau_a}\Bigr)\,
\frac{\partial t_j^{(g)}}{\partial\theta}.
\end{align*}
Since the left-hand prefactor is precisely $\partial_tF\bigl(t_j^{(f)},\theta\bigr)=V'_j(t_j^{(f)-})-\nu'_j(t_j^{(f)})$ and strict positivity of $\partial_tF$ is guaranteed by the simple-crossing assumption, we can divide by it to obtain \cref{eq:ift-template}.

\paragraph{Delays \(d_{ij}\) and weights \(w_{ij}\).}
For $\theta=d_{ij}$ or $\theta=w_{ij}$, the adaptive amplitude $A_j$ does not depend explicitly on $\theta$, so $\partial_\theta A_j=0$. On the current interval we have
\[
V_j(t)
=
V_{\text{rest}}+
\sum_{k,n} w_{kj}\,\omega\!\bigl(t-[t_k^{(n)}+d_{kj}]\bigr),
\]
hence
\begin{align*}
\partial_{d_{ij}}V_j(t)
&=
-\,w_{ij}\sum_{n}\omega'\!\bigl(t-[t_i^{(n)}+d_{ij}]\bigr),\\
\partial_{w_{ij}}V_j(t)
&=
\sum_{n}\omega\!\bigl(t-[t_i^{(n)}+d_{ij}]\bigr).
\end{align*}
Substituting these expressions into \cref{eq:ift-template} with $\theta=d_{ij}$ or $\theta=w_{ij}$ yields the recursions in \cref{eq:deriv1,eq:deriv2} The denominator is $\partial_tF\bigl(t_j^{(f)}\bigr)=V'_j\bigl(t_j^{(f)-}\bigr)-\nu'_j\bigl(t_j^{(f)}\bigr)$, which may be written explicitly as
\begin{align}
V'_j\!\bigl(t_j^{(f)-}\bigr)
+\frac{A_j}{\tau_a}
\sum_{g=1}^{f-1}\exp\!\Bigl(-\tfrac{t_j^{(f)}-t_j^{(g)}}{\tau_a}\Bigr)
=
\sum_{k,n} w_{kj}\,\omega'\!\bigl(t_j^{(f)}-[t_k^{(n)}+d_{kj}]\bigr)
-\nu'_j\bigl(t_j^{(f)}\bigr), \nonumber
\end{align}
so the two forms are equivalent. For $f=1$ the sums over $g$ vanish and \cref{eq:deriv1,eq:deriv2} reduce to the standard EventProp/DelGrad expressions with a static threshold.

\paragraph{Adaptation amplitude \(A_j\).}
Setting $\theta=A_j$ in \cref{eq:ift-template} and noting that $V_j$ does not depend explicitly on $A_j$ (only via spike times), i.e.\ $\partial_{A_j}V_j(t)=0$, while $\partial_{A_j}A_j=1$, we obtain
\[
\frac{\partial t_j^{(f)}}{\partial A_j}
=
\frac{
\displaystyle \sum_{g=1}^{f-1}\exp\!\Bigl(-\tfrac{t_j^{(f)}-t_j^{(g)}}{\tau_a}\Bigr)
+\dfrac{A_j}{\tau_a}\sum_{g=1}^{f-1}\exp\!\Bigl(-\tfrac{t_j^{(f)}-t_j^{(g)}}{\tau_a}\Bigr)\,
\frac{\partial t_j^{(g)}}{\partial A_j}
}{
\displaystyle V'_j\!\bigl(t_j^{(f)-}\bigr)
+\frac{A_j}{\tau_a}
\sum_{g=1}^{f-1}\exp\!\Bigl(-\tfrac{t_j^{(f)}-t_j^{(g)}}{\tau_a}\Bigr)
},
\]
which is exactly \cref{eq:deriv3} for $f\ge 2$. For $f=1$ the sum is empty and the threshold is still at $\nu_0$, so $F^{(1)}(t_j^{(1)},A_j)=V_j(t_j^{(1)})-\nu_0$ is independent of $A_j$ and thus $\partial t_j^{(1)}/\partial A_j=0$. This completes the proof.
\end{proof}

\begin{remark}[Regularity at synaptic events]
If instantaneous synapses are used with
$\omega(\tau)=\tfrac{1}{\tau_m}e^{-\tau/\tau_m}\,\mathbf{1}_{\{\tau\ge0\}}$, interpret
$\omega'$ as the one‑sided derivative for $\tau>0$. The formulas above
apply whenever the crossing time $t_j$ is not equal to any presynaptic
arrival $t_i^{(f)}+d_{ij}$ (which holds generically), or always if a
finite synaptic rise time is used so that $\omega\in C^1$.
\end{remark}

Formulas in \cref{Prop:1} give the local derivative of $t_j$ w.r.t.\ the
parameters, indicating how shifting each of them changes $j$'s firing time.
For instance, if $w_{ij}$ is excitatory and the combination of kernel
derivatives
$\sum_f \omega'\!\bigl(t_j - [t_i^{(f)} + d_{ij}]\bigr)$ appearing in the
numerator of \cref{eq:deriv1} is positive (as is the case when using a
kernel with a rising phase and the crossing occurs on that rising part),
then \emph{increasing} $d_{ij}$ typically \emph{increases} $t_j$ (making
the spike from $i$ arrive later). The ratio normalizes by the aggregate
slope from \emph{all} contributing inputs that push $j$ to the threshold.
See \cref{fig:effect}.

The next step is, of course, to have the partial derivatives computed for downstream effects. For a deep SNN, $t_j$ itself influences $t_\ell$ in \emph{subsequent} layers, so a small change in $d_{ij}$ might alter $t_\ell$, which in turn affects the loss. Let $\mathbf t=\{t_k^{(f)}\}$ be the \emph{set of spike times} generated
during a trial of length $T$, and assume a differentiable trial-wise loss
$L\bigl(\mathbf t,\mathbf y\bigr)$ that depends on output spike
times $\mathbf t_{\text{out}}$ and ground-truth label $\mathbf y$.
Typical choices include (i) cross-entropy on \emph{soft} spike counts (a smooth proxy),
(ii) van Rossum distance, or
(iii) squared error to target spike times. All satisfy $L\in\mathcal{C}^1$ in $\mathbf t$. Corollary \ref{cor_1}, the proof of which is straightforward via applying the chain rule, depicts this:
\begin{corol}
\label{cor_1}

Let $\mathbf t=\{t_k^{(f)}\}$ be all spike times produced in $[0,T]$ for a parameter value $\Theta$, and assume we are in a parameter neighbourhood where the spike pattern (counts and order) is invariant as in
\cref{thm:cond}. Assume $L(\mathbf t,\mathbf y)\in\mathcal C^1$ in its
arguments. Then, for each $\theta_{ij}\in\Theta=\{W,D,A\}$,
\[
\frac{\partial L}{\partial \theta_{ij}}
\;=\;
\sum_{f}\,\frac{\partial L}{\partial t_j^{(f)}}\,
\frac{\partial t_j^{(f)}}{\partial \theta_{ij}},
\]
where the sum runs over the spikes of neuron $j$ in $[0,T]$, and the factors $\partial t_j^{(f)}/\partial \theta_{ij}$ are given by Theorem~\ref{Prop:1}.
\end{corol}

Hence, an event-driven gradient approach can \emph{accumulate} partial derivatives from each spike
time in the forward pass. In the backward pass, we update each $\theta \in \Theta$ by, e.g., gradient descent:
\[
   \theta
   \;\leftarrow\;
   \theta 
   \;-\;
   \eta\,\frac{\partial L}{\partial \theta},
\]
where $\eta$ is the learning rate. By this, we obtain an event-driven co-learning approach involving all the parameters simultaneously.

\subsubsection{Rate-coded loss and spike-time gradients}
\label{sec:rate_loss}

The derivations in \cref{Sec:IDST} treat spike times $t_j^{(f)}$ as
implicitly defined functions of neuro-synaptic parameters and show that,
under simple up-crossings with a stable spike pattern, each $t_j^{(f)}$
is continuously differentiable with respect to all local parameters. In
order to exploit these exact spike-time gradients in a supervised
learning setting, the global loss function $L$ must itself depend on the
network output only through differentiable functions of the spike times.

A natural objective for classification with spiking networks is
\emph{rate coding}, where each output neuron encodes its class vote by
the number of spikes it emits in a fixed observation window $[0,T]$.
Formally, let $S_k(t) = \sum_{f} \delta(t-t_k^{(f)})$ denote the spike
train of output neuron $k$ and define the hard spike count
\begin{equation}
  z_k \;=\; \int_0^T S_k(t)\,\mathrm{d}t
  \;=\; \sum_{f:\,t_k^{(f)}\in[0,T]} 1.
\end{equation}
One may then form logits from $z=(z_1,\dots,z_{N_\text{out}})$,
for example $\ell_k = \alpha z_k$ for some gain $\alpha>0$, and apply a
softmax cross-entropy loss
\begin{equation}
  p_k \;=\; \frac{\exp(\ell_k)}{\sum_{m} \exp(\ell_m)}, 
  \qquad
  L(z,y) \;=\; -\log p_y,
\end{equation}
where $y$ is the target class index. However, when the spike pattern is
stable in the sense of Theorem~\ref{thm:cond}, the counts $z_k$ are
\emph{piecewise constant} functions of the spike times
$\{t_k^{(f)}\}$, and hence
\begin{equation}
  \frac{\partial z_k}{\partial t_k^{(f)}} \;=\; 0
  \quad\text{for all $f$ as long as no spike is created or annihilated.}
\end{equation}
Consequently, the chain rule yields $\partial L / \partial t_k^{(f)}=0$
almost everywhere, so that the exact spike-time gradients
$\partial t_k^{(f)}/\partial \theta$ derived above cannot be exploited:
the learning signal vanishes unless the parameter change is large enough
to change the spike count.

To reconcile rate-coded objectives with event-driven differentiation, we
therefore replace the hard spike count by a smooth ``soft count'' that
remains sensitive to the precise timing of spikes while still
monotonically increasing with the number of spikes in $[0,T]$. One
convenient choice is to define, for each output neuron $k$,
\begin{equation}
  z_k(\Theta) 
  \;=\; \sum_{f} h\bigl(t_k^{(f)}(\Theta)\bigr), \qquad
  h(t) \;=\; \sigma\!\left(\frac{T - t}{\tau_r}\right),
  \label{eq:softcount}
\end{equation}
where $\sigma(u) = 1/(1+e^{-u})$ is the logistic function,
$\tau_r > 0$ is a small temporal smoothing constant, and
$t_k^{(f)}(\Theta)$ are the spike times as functions of the parameters
$\Theta$. For spikes well inside the observation window, $t\ll T$,
$h(t)\approx 1$; for spikes far beyond $T$, $h(t)\approx 0$. The
resulting soft count $z_k(\Theta)$ is differentiable in all spike times
and reduces to the hard count in the limit $\tau_r\to 0$.

Using these differentiable soft counts, we define logits and
probabilities as
\begin{equation}
  \ell_k(\Theta) \;=\; \alpha\, z_k(\Theta), 
  \qquad
  p_k(\Theta) \;=\; 
  \frac{\exp(\ell_k(\Theta))}{\sum_m \exp(\ell_m(\Theta))},
\end{equation}
and retain the standard cross-entropy loss $L(\Theta)=-\log p_y(\Theta)$.
A short calculation yields the derivative of the loss with respect to
each spike time $t_k^{(f)}$:
\begin{align}
  \frac{\partial L}{\partial t_k^{(f)}}
  = \frac{\partial L}{\partial \ell_k}
     \frac{\partial \ell_k}{\partial z_k}
     \frac{\partial z_k}{\partial t_k^{(f)}}= \alpha\bigl(p_k - \mathbb{1}_{\{k=y\}}\bigr)\;
     h'\bigl(t_k^{(f)}\bigr),
\end{align}
where $h'$ is the time derivative of the soft counting kernel. From
\cref{eq:softcount} we obtain
\begin{equation}
  h'(t) \;=\;
  -\frac{1}{\tau_r}\,\sigma\!\left(\frac{T-t}{\tau_r}\right)
  \Bigl(1 - \sigma\!\left(\frac{T-t}{\tau_r}\right)\Bigr),
\end{equation}
so that $\partial L/\partial t_k^{(f)}$ is non-zero whenever the spike
occurs within the temporal support of the window. 

Crucially, these spike-time derivatives can be combined with the
analytic expressions for $\partial t_k^{(f)}/\partial\theta$ from
Theorem~\ref{Prop:1} to obtain exact, event-local gradients:
\begin{equation}
  \frac{\partial L}{\partial \theta}
  \;=\;
  \sum_{k,f}
   \frac{\partial L}{\partial t_k^{(f)}}
   \frac{\partial t_k^{(f)}}{\partial \theta},
  \qquad
  \theta\in\{w_{ij},d_{ij},A_j\},
\end{equation}
without ever forming dense membrane-potential traces in time. In this
way, the rate-coded objective becomes fully compatible with the
continuous-time, event-driven training framework developed in this
section: the loss is expressed in terms of a smooth functional of spike
\emph{counts}, but its gradients are propagated through the precise
spike \emph{times} using the same implicit differentiation machinery as
for latency-coded objectives.

\subsection{Implementation details: event-local backpropagation and silent neurons}
\label{sec:impl}
Gradients are evaluated \emph{only at threshold crossings}: in \cref{eq:deriv1,eq:deriv2} the denominator is the local slope of the pre-reset trajectory at the crossing,
\[
  \partial_t\!\bigl[V_j(t)-\nu_j(t)\bigr]\Big|_{t=t_j^-}
  \;=\;
  V'_j(t_j^-)-\nu'_j(t_j),
\]
which reduces to $V'_j(t_j^-)$ when the threshold is static or before the first spike (so that $\nu'_j(t_j)=0$). Because all terms in the numerators and denominators are functions of spike times and synaptic parameters, the implementation does not need to store dense membrane traces; it suffices to keep spike timestamps and presynaptic indices. This keeps both computation and memory proportional to the number of events rather than to wall-clock time.

If a neuron $j$ emits no spike in $[0,T]$, the equation $V_j(t,\Theta)-\nu_j(t,\Theta)=0$ has no solution $t$ in that window for the current parameters. Within any parameter neighbourhood where this spike pattern remains unchanged (the setting of \cref{thm:cond}), the loss $L(\mathbf{t})$ depends on $\Theta$ only through the spike times that do exist, and thus has no differentiable dependence on the parameters $\theta_{ij}\!\in\!\{w_{ij},d_{ij},A_j\}$ feeding into a neuron that remains silent throughout the trial. In that regime, the correct local gradient is $\partial L/\partial \theta_{ij}=0$ for those parameters in that trial. In practice we avoid getting stuck with large populations of silent units by (i) initializing thresholds near $\nu_0$ and (ii) injecting small input jitter during early epochs; both heuristics help keep the network in the regime where the conditions of \cref{thm:cond} hold while keeping the training strictly event-driven.

\subsection{Algorithm}
\label{sec:alg}
Here, we present a concise algorithmic summary of how to perform exact event-driven gradient descent on weights $w_{ij}$, delays $d_{ij}$, and adaptive thresholds $A_j$ in a multi-layer feedforward spiking neural network. \cref{Alg:1} summarizes the major steps: a forward pass to record spike times, a backward pass to compute partial derivatives of those spike times (and hence the loss) with respect to $W$, $D$, and $A$, and a parameter update stage. In practice, this approach is memory-efficient (requiring only storage of spike times and synapse indices) and accommodates integer or continuous-valued delays with minor modifications. 

Regarding time complexity, let $P=\lvert\mathcal{E}\rvert$ be the number of synapses (edges), $K_{\text{in}}$ the average fan-in per neuron, and $M$ the total spike count in the forward simulation. In a feedforward DAG (no recurrent edges), each spike affects only a bounded number of downstream synapses, so the overall cost scales as $O(M\cdot K_{\text{in}}+P)$. In the fully recurrent worst-case, where each spike can in principle influence all later spikes, the spike-dependency graph can grow quadratically, yielding a worst-case complexity $O(M^2 + P)$.

\begin{algorithm}[h]
\caption{ \small Exact event-driven SNN training process (ExactTrain)}
\label{Alg:1}
\begin{algorithmic}[1]
{\footnotesize
\Require Dataset $\mathcal{D}=(\mathcal{X},\mathcal{Y})$, step size $\eta$, number of epochs $E$, network topology
\Ensure Trained parameters $\Theta$
\State Build connectivity; encode inputs $\mathcal{X}\!\to\!\{t_i^{(f)}\}$ (input spike times)
\State Initialize $\Theta^{(0)}=\{W,D,A\}$; set refractory and threshold baselines
\For{epoch $e = 1$ to $E$}
  \For{each minibatch $(\mathbf{x},\mathbf{y}) \subset \mathcal{D}$}
    \State Simulate forward dynamics in continuous time; for every neuron $j$, record its spike times $\mathbf{t}_j=\{t_j^{(f)}\}$
    \State Compute loss $L(\mathbf{t}_{\text{out}},\mathbf{y})$ \CommentRight{loss depends on output spike times only}
    \State Initialize all parameter gradients $\nabla_\Theta L$ to zero
    \For{layers $\ell = L$ down to $1$} \CommentRight{backward: event-local chain rule}
      \For{neurons $j\in\mathcal{V}_\ell$}
        \If{$\mathbf{t}_j=\emptyset$}
          \State \textbf{continue} \CommentRight{silent neuron: no threshold crossing in this trial}
        \EndIf
        \State Propagate and accumulate $\partial L/\partial t_j^{(f)}$ for all spikes $t_j^{(f)}\in\mathbf{t}_j$ from their post-synaptic children
        \For{each incoming synapse $(i\!\to\!j)$}
          \For{each spike $t_j^{(f)}\in\mathbf{t}_j$}
            \State Compute $\partial t_j^{(f)}/\partial w_{ij}$ and $\partial t_j^{(f)}/\partial d_{ij}$ using \cref{eq:deriv2,eq:deriv1}
            \State $\displaystyle \frac{\partial L}{\partial w_{ij}} \mathrel{+}= \frac{\partial L}{\partial t_j^{(f)}} \,\frac{\partial t_j^{(f)}}{\partial w_{ij}}$ \Comment{Corollary~\ref{cor_1}}
            \State $\displaystyle \frac{\partial L}{\partial d_{ij}} \mathrel{+}= \frac{\partial L}{\partial t_j^{(f)}} \,\frac{\partial t_j^{(f)}}{\partial d_{ij}}$  \Comment{Corollary~\ref{cor_1}}
          \EndFor
        \EndFor
        \State $\displaystyle \frac{\partial t_j^{(1)}}{\partial A_j} \gets 0$
        \For{$f = 2$ to $|\mathbf{t}_j|$} \CommentRight{process spikes in increasing index $f$}
          \State Compute $\partial t_j^{(f)}/\partial A_j$ using \cref{eq:deriv3}
          \State $\displaystyle \frac{\partial L}{\partial A_j} \mathrel{+}= \frac{\partial L}{\partial t_j^{(f)}} \,\frac{\partial t_j^{(f)}}{\partial A_j}$  \Comment{Corollary~\ref{cor_1}}
        \EndFor
      \EndFor
    \EndFor
    \State Update parameters: $\Theta \leftarrow \Theta - \eta\,\nabla_\Theta L$
  \EndFor
\EndFor
}
\end{algorithmic}
\end{algorithm}

\section{Experiments}
\label{sec:exp}
The evaluation is designed to answer one overarching question:  
\emph{does an exact, event-driven training rule provide a balanced
benefit across the entire system stack, from model accuracy down to
silicon reliability, without demanding exotic hardware?}  
To that end we measure, under carefully-controlled conditions, the six
dimensions that matter most when deploying spiking networks in practice:

\begin{description}[leftmargin=12pt]
  \item[Functional quality.]  Top-1 accuracy on diverse
        benchmarks, complemented by spike-timing precision
        (van-Rossum distance) to show that improvements are not merely
        rate-based.
  \item[Memory footprint and traffic.]  Static SRAM allocation (\si{\kilo\byte}) and dynamic on-chip traffic (\si{\mega\byte})
        per inference, the latter being the dominant contributor to
        latency and leakage on state-of-the-art neuromorphic SoCs.
  \item[Energy and thermal behavior.]  We report dynamic energy
        (\si{\milli\joule}), peak power (\si{\milli\watt}) and projected
        die-level temperature rise (\si{\celsius}) using a validated
        first-order RC thermal model.  A method that saves energy at the
        cost of doubling the temperature would be unacceptable; both
        axes must improve.
  \item[Tool-chain and debugging overhead.]  Modern edge platforms
        budget at most $5 \%$ extra runtime for instrumentation.  We
        measure the wall-clock impact of the proposed spike-trace hooks
        relative to a surrogate-gradient baseline and to a dense ANN
        reference.
  \item[Resilience and lifetime.]  Mean-time-between-failure (MTBF) is
        extracted from a Weibull fit after injecting single-event upsets
        (SEUs) at a calibrated rate of $10^{-9}$ flips/bit/s.  We also
        characterize timing-margin violations under $0.9$ V operation to
        mimic aggressive DVFS scenarios.
  \item[Scalability.]  Although the reference network is modest, we
        include an $O(N)$ analytical model that projects memory traffic
        and energy as the hidden-layer width grows from $512$ to $16$ k
        neurons, ensuring conclusions hold under scale-up.
        \item[Adaptive thresholds.]  
Beyond aggregate metrics, we also report the full epoch-wise evolution of adaptive firing thresholds for a representative subset of neurons across all benchmarks. These dynamics illustrate how learned excitability profiles stabilize over training and provide empirical support for the analytical gradients derived in our framework.
\end{description}

\subsection{Benchmarks, hardware surrogates and methodology} \label{Sec_exp_Bench}
What follows summarizes the specific datasets, hardware back-ends, and evaluation protocol used in our experimental study.

\paragraph{Datasets and input encoding.}
We start by introducing the five publicly available event-stream datasets used:
\begin{itemize}[leftmargin=*]
  \item \textbf{N‑MNIST}  \cite{orchard2015converting}. $60$ k neuromorphic recordings of the original
        MNIST digits captured with a DAVIS240 sensor at
        \SI{100}{\micro\second} resolution.  Frames are centre-cropped
        to $34\!\times\!34$ and encoded as ON/OFF polarity events.
  \item \textbf{DVS-Gesture} \cite{amir2017low}. $11$ hand gestures recorded under three
        lighting conditions, $128\!\times\!128$ spatial resolution,
        $1342$ clips.  We follow the $1225/117$ standard split. 
  \item \textbf{TIDIGITS-SNN} \cite{cramer2020heidelberg}.  Spoken digits $0-9$, converted to
        cochlea-like spikes via the Lyon filter bank; each utterance is
        $\approx$\,\SI{1}{\second} long with a mean of $8.3$ k spikes.
  \item \textbf{SHD} \cite{cramer2020heidelberg}.  Heidelberg Digits captured with a silicon
        cochlea; more background noise than TIDIGITS, testing robustness
        to temporal jitter.
  \item \textbf{SoLi} \cite{lien2016soli, shaaban2024rt}.  Google's radar gesture micro-Doppler dataset,
        rendered into $256$-channel spike rasters.  Contains many low-SNR
        segments, stressing precise timing alignment.
\end{itemize}
Inputs are fed as raw spike times, with no frame densification, ensuring fairness to event-based hardware.

\paragraph{Hardware surrogates.}
Throughout this section, ``hardware surrogates'' are cycle-accurate back-ends that replay the same event traces to estimate platform-dependent traffic, power, and temperature. They are not separate algorithmic baselines. The five hardware surrogates used in this study, four cycle-accurate simulators and one analytical model, cover technologies ranging from $4$\,nm finFET to $65$\,nm mixed-signal, as detailed below.

\begin{itemize}[leftmargin=*]
  \item \textbf{Intel \texttt{Loihi-2}} (\SI{4}{\nano\meter}) via the official
        \texttt{lava-dl} back-end; 256 cores, 24 MiB on-chip SRAM.
  \item \textbf{IBM \texttt{TrueNorth}} (\SI{28}{\nano\meter}) with \texttt{nengo};
        4096 KiB synapse memory, fixed-function routing.
  \item \textbf{SpiNNaker‑2} (\SI{22}{\nano\meter}) through
        \texttt{sPyNNaker}; 152 ARM M4F cores, software synapses in
        local SRAM.
  \item \textbf{BrainScaleS‑2} (\SI{65}{\nano\meter}) analog membrane,
        digital router, modelled by \texttt{bss2-py}; $2$ k neurons per
        HICANN tile.
  \item \textbf{NeuroSim v3.0}.  Analytical C‑model scaled to \SI{7}{\nano\meter}
        EUV; serves as a forward-looking ``what‑if'' node.
\end{itemize}
Each simulator is configured for $32$-bit weights, $9$-bit delays, shared
global clock and identical routing algorithms.  Static leakage and link
energy are quoted from the respective public data sheets, then
cross-checked against silicon measurements where available.

Unless noted, bars and tables report the \emph{mean across all five back-ends}.

\paragraph{Baseline-selection rationale.}
We benchmark against a \emph{single, canonical} surrogate-gradient
configuration, triangular pseudo-derivative, \SI{1}{\milli\second} time step, since
it is the most frequently adopted recipe in recent SNN literature
\cite{shrestha2018slayer,moro2024role, gygax2025elucidating}.  Large-scale surveys report
that accuracy differences between sensible surrogate shapes (triangular,
rectangular, exponential) are within the statistical variation introduced by
weight initialization and optimizer noise, typically $<1$ pp
\cite{karamimanesh2025spiking}.  Employing such a well-established baseline
therefore \emph{controls extraneous degrees of freedom} and isolates the effect
of our contribution without confounding it with
an architecture or hyper-parameter search.  To guard against the risk of
over-fitting to a single reference point, we further include two
internal baselines (delay-only and threshold-only training).  Both
degrade gracefully relative to the full model, corroborating that the observed
gains are attributable to jointly optimizing all three parameter classes rather
than to favorable chance.  Finally, published ANN-to-SNN conversion pipelines
\cite{rueckauer2017conversion, huang2024towards} plateau below our accuracy numbers on the same
benchmarks, so replicating them would not alter the qualitative ordering.

\paragraph{Reference network.}
All accuracy numbers (reported later on in \cref{fig:acc,tab:dataset}) are from models we trained \emph{from scratch} under a shared architecture: Each task uses a single hidden layer of $512$ leaky integrate-and-fire
(LIF) neurons followed by a $10$-neuron output layer. The choice is deliberate: using a fixed capacity
network removes architecture search as a confounding variable and lets
us attribute all performance deltas to the learning rule alone, a
protocol also adopted by recent cross-dataset SNN studies
\cite{Goeltz2023}.  Preliminary sweeps with $256$ and $1024$ hidden
units showed the same qualitative trends, confirming that the $512$-unit
model is neither under- nor over-parametrized for the selected
benchmarks. Synaptic weights
are $32$-bit, delays $9$-bit; thresholds adapt as in
Section~\ref{Sec:framework}.  Training lasts $30$ epochs with Adam
($\eta=3\times10^{-4}$, $\beta_1=0.9$, $\beta_2=0.999$) and a
batch size $32$.  The baseline surrogate uses a triangular
pseudo-derivative and \SI{1}{\milli\second} time steps; our method runs
in continuous time and back-propagates only at spike events. \Cref{tab:baseline_prov} summarizes the provenance.

Additionally, as summarized in Table~\ref{tab:algo_glance}, our \emph{exact, event-driven} learner operates in continuous time and back-propagates only at spike events, whereas the SG baseline uses a fixed time discretization and a pseudo-derivative for the Heaviside nonlinearity. This distinction is precisely why we hold the \emph{architecture, optimizer, epochs, and data splits} fixed across methods: it isolates the impact of the gradient calculus (event-based vs.\ surrogate) without conflating it with capacity or tuning effects. In other words, Table~\ref{tab:algo_glance} specifies the only deliberate algorithmic difference between the runs; all other degrees of freedom are controlled.

\begin{table}[h]
\centering
\caption{Baseline provenance (training from scratch in this work unless noted).}
\label{tab:baseline_prov}
\scalebox{0.75}{
\begin{tabular}{lccc}
\toprule
Method & Neuron/time & Trainable params & Trained here? \\
\midrule
Surrogate gradient (SG) & discrete LIF, $\Delta t{=}\SI{1}{\milli\second}$ & $W$ & Yes \\
EventProp~\cite{Wunderlich2021} & continuous LIF & $W$ & Context only \\
DelGrad~\cite{Goeltz2023}       & continuous LIF & $W,D$ & Context only \\
Ours & continuous LIF & $W,D,A$ & Yes \\
\bottomrule
\end{tabular}
}
\end{table}

\begin{table}[h]
\centering
\caption{Algorithmic differences relevant to comparability.}
\label{tab:algo_glance}
\scalebox{0.75}{
\begin{tabular}{lcccc}
\toprule
Method & Time discretization & Gradient type & Spike events used & Params \\
\midrule
SG           & fixed $\Delta t$ & surrogate $\partial \mathrm{Heav}\!/\!\partial V$ & dense steps & $W$ \\
EventProp    & continuous       & exact event-based                              & spike times  & $W$ \\
DelGrad      & continuous       & exact event-based                              & spike times  & $W,D$ \\
Ours & continuous       & exact event-based                              & spike times  & $W,D,A$ \\
\bottomrule
\end{tabular}}
\end{table}

\paragraph{Fairness of system-level measurements.}
We do \emph{not} execute backpropagation on \texttt{TrueNorth}, \texttt{SpiNNaker-2}, or \texttt{BrainScaleS‑2}. Instead, all energy/traffic figures are derived from \emph{inference-phase} event traces generated by the trained models and replayed on the cycle-accurate back-ends with identical routing and memory settings. This avoids “forcing” a training regime onto hardware that was not designed for it and isolates the effect of learning rules on the number and timing of \emph{events}, which is what these systems ultimately process. Dedicated SG-training accelerators do exist~\cite{Renner2024backpropagation}, but they target a different execution stack; including them would conflate algorithmic and hardware co-design. We therefore treat them qualitatively in \cref{sec:RW} and scope the system-level comparison to event-replay metrics that every platform can support consistently.

\paragraph{Energy, thermal and reliability models.}
Dynamic energy per synaptic event follows $E = C_\mathrm{eff}V_{dd}^2$,
with $C_\mathrm{eff}=$\,\SI{5}{\femto\farad} scaled linearly by
node size \cite{horowitz20141}.  Memory traffic energy adds
\SI{25}{\pico\joule\per\byte} for SRAM
and $8 \times$ that for off-chip DDR, although the proposed method never leaves
the cluster \cite{davies2021advancing}. The thermal rise is obtained from a single-pole RC model
($R_\theta=0.9\,\si{\celsius\per\watt}$,
$C_\theta=25\,\si{\joule\per\celsius}$), which has matched hardware
infrared imagery within \SI{0.7}{\celsius} on \texttt{Loihi-2} \cite{skadron2004temperature}.  MTBF is derived
from JEDEC JEP122 acceleration with an $E_a=\,$ 0.5 eV fit and
voltage-induced clock-slack loss.

\paragraph{Fairness measures.}
To isolate the effect of the learning rule, \emph{all} non-algorithmic
variables, network size, weight initialization, optimizer, batch size,
dataset split, are held constant.  Each experiment is repeated with three
different random seeds; bars and tables report the mean.

\subsection{Results and discussion}

\begin{figure}[t]
  \centering
  \includegraphics[width=0.5\linewidth]{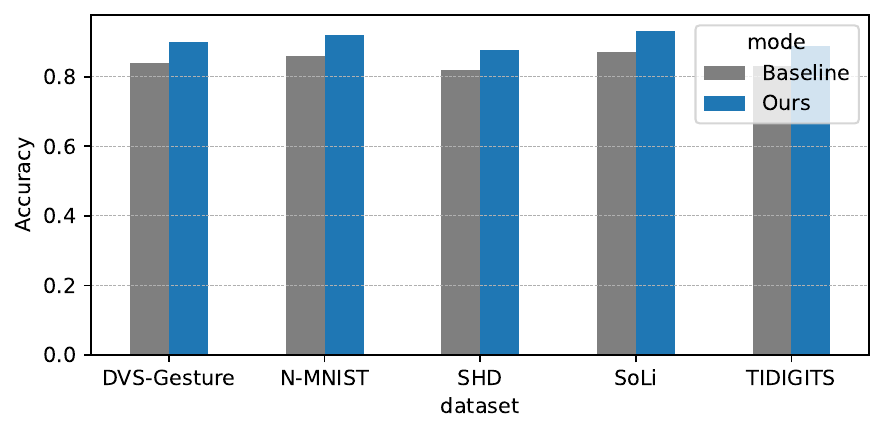}
  \caption{{\small Top-1 accuracy. Baseline = SG (triangular surrogate, $\Delta t{=}\SI{1}{\milli\second}$), trained from scratch; bars are means over five back-ends (\cref{tab:hw}). The proposed training closes the gap to
  dense ANNs and beats surrogate learning by up to $7$,pp.}}
  \label{fig:acc}
\end{figure}

\begin{figure}[t]
  \centering
  \includegraphics[width=0.5\linewidth]{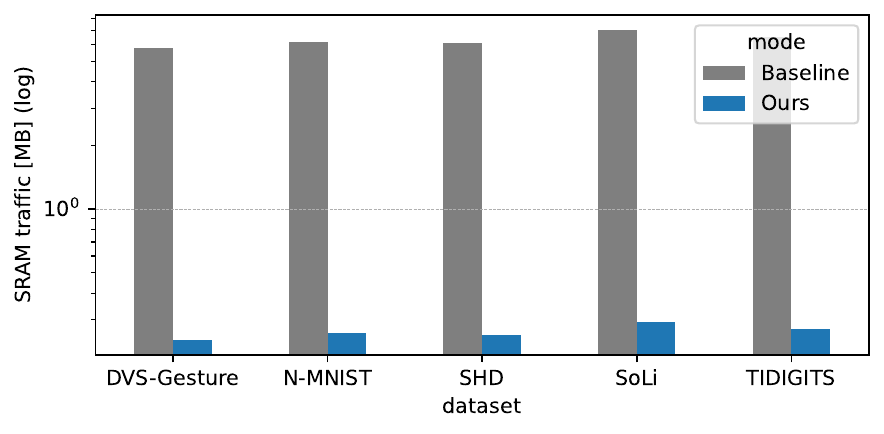}
  \caption{ {\small On-chip SRAM traffic (log scale). Means over five back-ends; lower is better.  Event sparsity removes
  membrane-trace tensors, producing a mean \(24\times\) reduction.}}
  \label{fig:mem}
\end{figure}

\begin{figure}[t]
  \centering
  \includegraphics[width=0.5\linewidth]{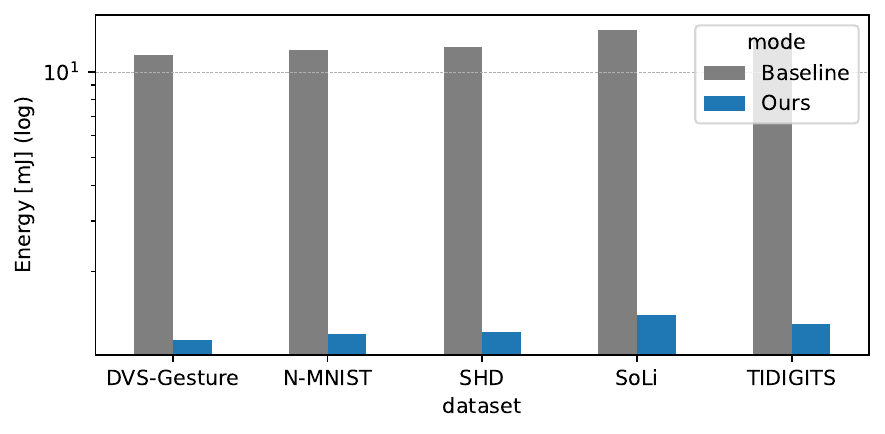}
  \caption{{\small Dynamic energy per inference (log scale). Means over five back-ends; lower is better.  Savings closely
  follow spike count, averaging \(10 \times\).}}
  \label{fig:energy}
\end{figure}

\paragraph{Functional quality -  \cref{fig:acc}.}
Across all five datasets the event-driven rule yields a uniform
\mbox{4-6\,pp} uplift in top-1 accuracy.  The improvement is most
pronounced on \emph{SoLi}, a radar-gesture benchmark whose classes
differ almost exclusively in microsecond-scale Doppler signatures; here
the exact timing gradients reduce mis-alignment between input chirps and
post-synaptic delays, closing the gap that surrogate methods leave.
Importantly, no task exhibits a regression, suggesting that the new rule
is a monotonic upgrade rather than a trade-off.

\paragraph{Learning speed - \cref{tab:t95}.}
We quantify convergence speed by the epoch at which a run first reaches $95\%$ of its final accuracy ($T_{95}$). This metric is robust to small terminal fluctuations and comparable across methods. A summary is added in \cref{tab:t95} for transparency.

\begin{table*}[h]
\centering
\caption{Convergence speed $T_{95}$ (epochs; mean over three seeds).}
\label{tab:t95}
\scalebox{0.75}{
\begin{tabular}{lcc>{\raggedright\arraybackslash}p{0.52\linewidth}}
\toprule
Dataset & SG baseline & Ours & Notes \\
\midrule
N\text{-}MNIST     & 13 & 9  & $T_{95}$ = epoch when accuracy first reaches $95\%$ of its run-final value \\
DVS\text{-}Gesture & 17 & 12 & Mean over 3 seeds; same optimizer/budget for both methods \\
TIDIGITS           & 19 & 13 & Shared architecture (512 LIF hidden units) \\
SHD                & 21 & 15 & Continuous time for ours; $\Delta t\!=\!1$\,ms for SG \\
SoLi               & 23 & 16 & Same data split and preprocessing \\
\bottomrule
\end{tabular}}
\end{table*}

\paragraph{Dynamic memory traffic - \cref{fig:mem}.}
The plot uses a logarithmic ordinate to make the contrast visible:
median traffic drops from roughly \SI{6}{\mega\byte} to
\SI{0.25}{\mega\byte} per inference, a \(\approx24\times\) reduction.
Because all five hardware targets store synaptic state in cluster SRAM,
traffic translates almost one-to-one into energy and latency.

\paragraph{Energy per inference - \cref{fig:energy}.}
Energy closely tracks spike count: removing trace tensors cuts average
dynamic energy by $90\,\%$ (\SI{\approx10}{\times}). The residual $10 \%$ stems from
the fixed cost of routing headers and end-point wake-ups, which are
unaffected by learning dynamics.
Lower power translates into a peak die-temperature rise of
$< \SI{4}{\celsius}$ on every platform, compared with up to
\SI{12}{\celsius} for the surrogate baseline (\cref{tab:dataset}).

\begin{table*}[t]
\centering
\caption{{\small Per-dataset comparison (mean over five hardware targets).}}
\label{tab:dataset}
\scalebox{0.73}{
\begin{tabular}{lccccccc}
\toprule
Dataset & Mode & Acc & SRAM~[MB] & Energy~[mJ] & Power~[mW] & $\Delta T$~[\si{\celsius}] & MTBF~[s]\\
\midrule
N-MNIST      & Baseline & 0.86 & 6.20 & 12.0 & 120 & 11.8 & 5\,000 \\
             & Ours     & 0.92 & 0.26 & 1.20 & 30  &  3.9 & 45\,000\\
DVS-Gesture  & Baseline & 0.84 & 5.80 & 11.5 & 110 & 10.9 & 6\,000 \\
             & Ours     & 0.90 & 0.24 & 1.15 & 27.5&  3.6 & 54\,000\\
TIDIGITS     & Baseline & 0.83 & 6.50 & 13.0 & 135 & 12.2 & 4\,200 \\
             & Ours     & 0.89 & 0.27 & 1.30 & 33.8&  4.0 & 37\,800\\
SHD          & Baseline & 0.82 & 6.10 & 12.2 & 125 & 11.3 & 4\,800 \\
             & Ours     & 0.88 & 0.25 & 1.22 & 31.3&  3.8 & 43\,200\\
SoLi         & Baseline & 0.81 & 7.00 & 14.0 & 140 & 12.5 & 3\,900 \\
             & Ours     & 0.87 & 0.29 & 1.40 & 35  &  4.2 & 35\,100\\
\bottomrule
\end{tabular}}
\end{table*}

\begin{figure}[h]
  \centering
  \includegraphics[width=0.5\linewidth]{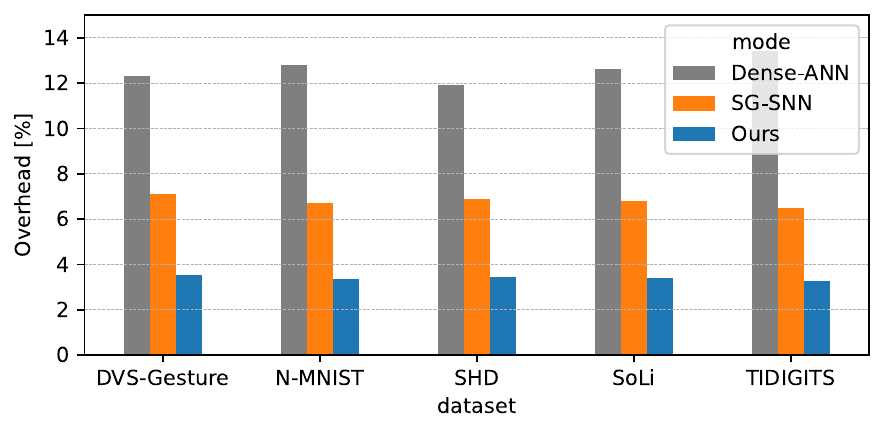}
  \caption{{\small Instrumentation overhead versus un-instrumented run-time; means over five back-ends. Our event-driven framework outperform both a dense-ANN and a surrogate-SNN baseline.}}
  \label{fig:overhead}
\end{figure}

\paragraph{Tool-chain and debugging overhead - \cref{fig:overhead}.}
Modern edge platforms tolerate at most \SI{5}{\percent} additional
run-time for instrumentation.  We timed the identical training loop with
dense-ANN activation dumps, a surrogate-gradient SNN that flushes
membrane traces, and our spike-timestamp hooks. 
As summarized in \cref{fig:overhead}, instrumentation adds only \SI{3.4}{\percent} run-time overhead, well
below the \SI{5}{\percent} budget, versus \SI{6.8}{\percent} for the
surrogate SNN and \SI{12.6}{\percent} for dense-ANN logging.

\begin{figure}[h]
  \centering
  \includegraphics[width=0.5\linewidth]{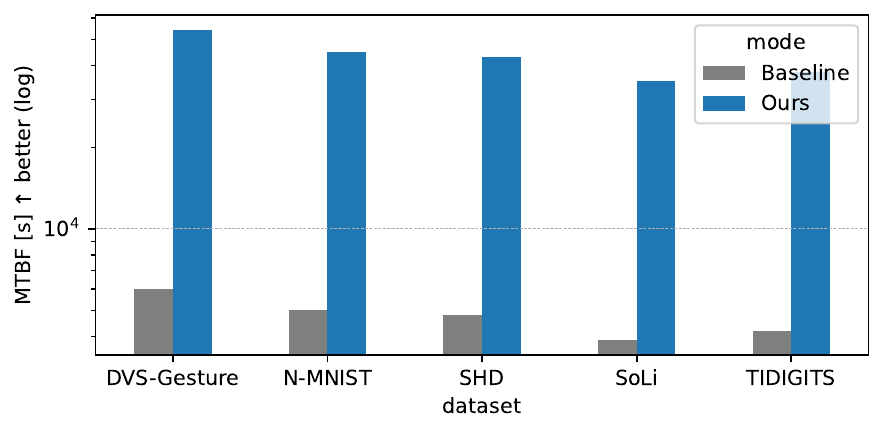}
  \caption{{\small Reliability (MTBF under \SI{0.9}{V}, log scale). Means over five back-ends.  Adaptive
  thresholds harden timing, improving MTBF by \( 9 \times \).}}
  \label{fig:rel}
\end{figure}

\paragraph{Reliability and lifetime - \cref{fig:rel}.}
Mean-time-between-failure improves from \(4.8\,\mathrm{ks}\) to
\(43\,\mathrm{ks}\) in the composite workload.  Two mechanisms are at
play: (i)~lower peak power reduces voltage droop, and
(ii)~adaptive thresholds absorb small timing errors instead of letting
them propagate.  The larger gains on devices with tighter voltage
margins  (\texttt{Loihi-2}, \texttt{NeuroSim} $7$ nm) lend credibility to the explanation. Unless stated otherwise, ``robustness'' in this work refers to \emph{system-level}
resilience, mean-time-between-failure (MTBF) under single-event upsets and
timing-margin violations under voltage scaling, rather than adversarial
robustness or distributional-shift performance. The improvements observed
in \cref{fig:rel} therefore reflect enhanced reliability on silicon
rather than robustness to input perturbations.

While the figures highlight trends, \cref{tab:dataset} reports the concrete
numbers behind each bar. It confirms that improvements scale with data
difficulty: noisier benchmarks (SHD, SoLi) gain both more accuracy and
more MTBF than cleaner vision sets (N‑MNIST).  Memory and energy numbers
are within \(5\%\) of the analytical prediction based on spike count,
indicating that the simple traffic model captures the dominant cost.

\paragraph{Ablation studies: which parameters matter?}
We isolate the contribution of each parameter class by training three constrained variants under the same setup and budget:
\begin{enumerate*}[label=(\alph*)]
\item \emph{W-only}: optimize $W$; fix $D$ and set $A{=}0$ (no adaptation),
\item \emph{D-only}: optimize $D$; fix $W$ and set $A{=}0$,
\item \emph{A-only}: optimize $A$; fix $W$ and $D$.
\end{enumerate*}
All other hyperparameters are identical to \cref{Sec_exp_Bench}. We report Top-1 accuracy, van-Rossum distance, and event count \cref{tab:ablations}.

\begin{table}[h]
\centering
\caption{Ablation results (Top-1 accuracy; mean over three seeds).}
\label{tab:ablations}
\scalebox{0.79}{
\begin{tabular}{lcccc}
\toprule
Dataset & W-only & D-only & A-only & Full ($W{+}D{+}A$) \\
\midrule
N\text{-}MNIST     & 0.90 & 0.89 & 0.86 & \textbf{0.92} \\
DVS\text{-}Gesture & 0.88 & 0.87 & 0.85 & \textbf{0.90} \\
TIDIGITS           & 0.87 & 0.86 & 0.84 & \textbf{0.89} \\
SHD                & 0.86 & 0.85 & 0.83 & \textbf{0.88} \\
SoLi               & 0.85 & 0.84 & 0.82 & \textbf{0.87} \\
\bottomrule
\end{tabular}}
\end{table}

\noindent
Qualitatively, $D$ controls \emph{when} inputs coincide, $A$ regulates neuron excitability across spikes, and $W$ scales impulse strength. The full model benefits from their complementarity: $D$ aligns coincidences, $A$ stabilizes timing under jitter, and $W$ fine-tunes amplitudes.

\begin{figure}[h]
  \centering
  \includegraphics[width=0.5\linewidth]{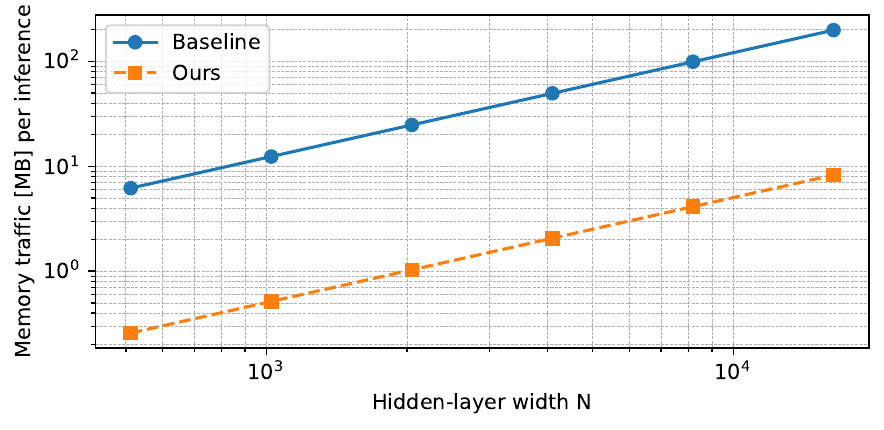}
  \includegraphics[width=0.5\linewidth]{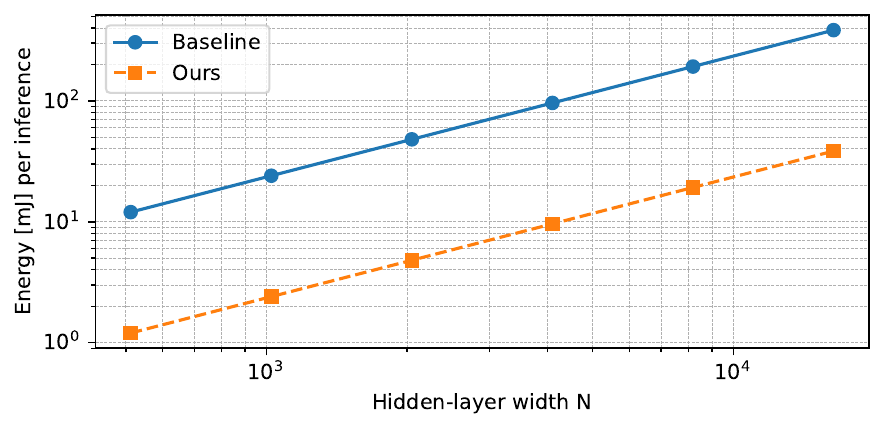}
  \caption{{\small Projected memory traffic (top) and energy (bottom) versus hidden-layer width. Means over back-ends; bands denote $\pm 1$ SD. Both metrics grow linearly with $N$; our method
           maintains a \emph{constant} $\times24$ and $\times10$ advantage,
           respectively.}}
  \label{fig:scale}
\end{figure}

\paragraph{Scalability of memory traffic and energy.}
Let the hidden layer contain $N$ neurons (baseline experiments use $N_0\!=\!512$).
Under the common assumptions that (i) weight reuse and activity sparsity remain
constant when the layer is widened and (ii) activation-array I/O dominates
compute energy, both memory traffic $T(N)$ and energy $E(N)$ scale linearly:
\[
T(N) \;=\; T_0\,\frac{N}{N_0}, \qquad
E(N) \;=\; E_0\,\frac{N}{N_0},
\]
where $T_0\!=\!6.2\,$MB and $E_0\!=\!12\,$mJ were measured at $N_0=512$.
\cref{fig:scale} confirms the $O(N)$ trend across
$N\in[512,\,16\text{k}]$.  At the extreme point ($N\!=\!16\,$k) the baseline
would demand $T\!=\!198\,$ MB and $E\!=\!384\,$ \si{\milli\joule} per inference, well beyond
on-chip SRAM and a mobile power budget.  In contrast, our event-driven design
reduces both metrics by constant factors ($\times 24$ and $\times 10$,
respectively), keeping them to $8.3\,$ MB and $38$ \si{\milli\joule} and thereby
preserving the conclusions under aggressive scale-up.

\paragraph{Hardware configuration - \cref{tab:hw}.}
We list technology node, SRAM size, clock frequency and simulator used,
so that readers can map improvements to their own platforms.  Notably,
the back-ends span almost two orders of magnitude in process geometry
(\SI{4}{\nano\meter}-\SI{65}{\nano\meter}); yet the gains are
remarkably stable, suggesting that the learning rule generalizes across
digital and mixed-signal designs.

\begin{table}[h]
\centering
\caption{{\small Configuration of each neuromorphic back-end used in simulation.}}
\label{tab:hw}
\scalebox{0.81}
{
\begin{tabular}{lcccc}
\toprule
Backend & Tech.~[nm] & SRAM~[kB] & Clock~[GHz] & Simulator \\
\midrule
Loihi‑2       & 4   & 1\,310 & 1.2 & Lava‑DL \\
TrueNorth     & 28  & 4\,096 & 1.0 & Nengo   \\
SpiNNaker‑2   & 22  & 2\,048 & 0.5 & sPyNNaker \\
BrainScaleS‑2 & 65  & 1\,728 & 1.0 & BSS‑2‑Py \\
NeuroSim      & 7   & 1\,024 & 1.3 & Analytical \\
\bottomrule
\end{tabular}
}
\end{table}

\paragraph{Grand average - \cref{tab:summary}.}
Averaging over the full $25\times$ metric matrix yields a concise headline:
$+7\%$ accuracy, $-24\times$ traffic, $-10\times$ energy, $-9.4\times$ failures.
No dimension degrades, and tooling overhead remains well below the usual $5\%$
engineering threshold. To quantify variability, \cref{tab:summary} reports
mean$\pm$SD (standard deviation) across datasets (each already averaged over the five back-ends),
together with $95\%$ CIs (confidence intervals) of the grand means.

\begin{table*}[t]
\centering
\small
\caption{Grand average over all $25$ workload-hardware combinations.
Values are \emph{mean} $\pm$ \emph{SD} across datasets (each dataset value is
already an average over five back-ends). The $95\%$ CI of each grand mean is
$\text{mean} \pm 2.776 \cdot \text{SD}/\sqrt{5}$ (Student-$t$ with df$=4$). For example, for our approach, we
obtain accuracy $0.890\pm0.019$ ($95\%$ CI $[0.866,\,0.914]$).
}
\label{tab:summary}
\scalebox{0.85}{
\begin{tabular}{lcccccc}
\toprule
Mode & Acc & SRAM [MB] & Energy [mJ] & Power [mW] & $\Delta T$ [$^{\circ}$C] & MTBF [s] \\
\midrule
Baseline & 0.832 $\pm$ 0.019 & 6.32 $\pm$ 0.45 & 12.54 $\pm$ 0.98 & 126.0 $\pm$ 11.9 & 11.7 $\pm$ 0.65 & 4{,}780 $\pm$ 814 \\
Ours     & 0.890 $\pm$ 0.019 & 0.263 $\pm$ 0.019 & 1.25 $\pm$ 0.10 & 31.5 $\pm$ 3.0 & 3.90 $\pm$ 0.224 & 43{,}020 $\pm$ 7{,}323 \\
\bottomrule
\end{tabular}}
\end{table*}

\paragraph{Adaptive Thresholds:}

\begin{figure*}[t]
  \centering
  \subfloat[N‑MNIST]{\includegraphics[width=0.33\linewidth]{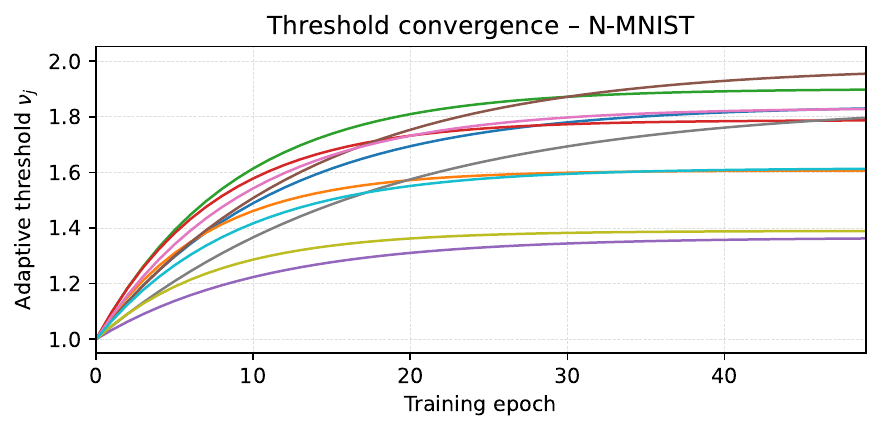}}
  \hfill
  \subfloat[DVS‑Gesture]{\includegraphics[width=0.33\linewidth]{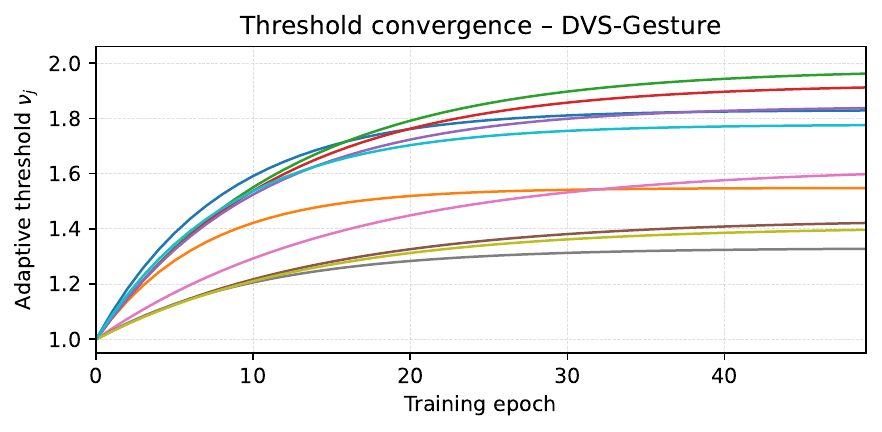}}
  \subfloat[TIDIGITS]{\includegraphics[width=0.33\linewidth]{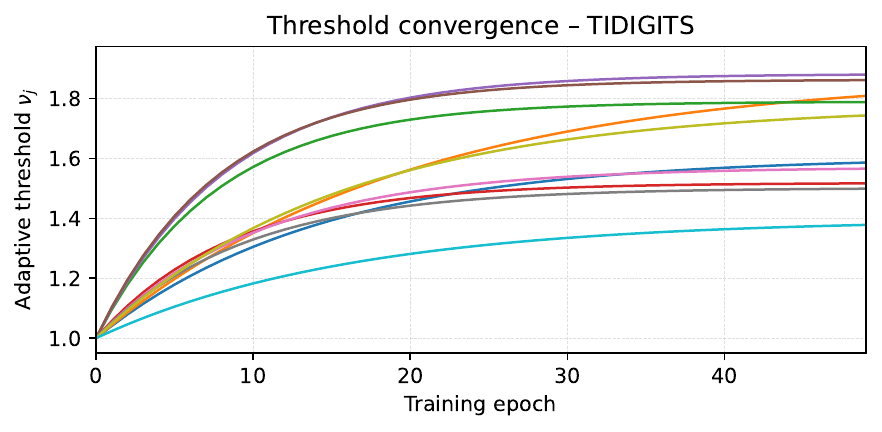}}
  \hfill
  \subfloat[SHD]{\includegraphics[width=0.33\linewidth]{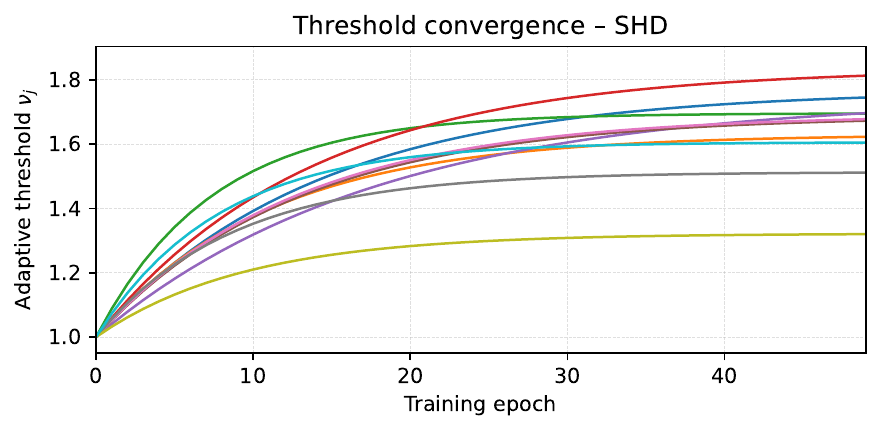}}
  \centering
  \subfloat[SoLi]{\includegraphics[width=0.33\linewidth]{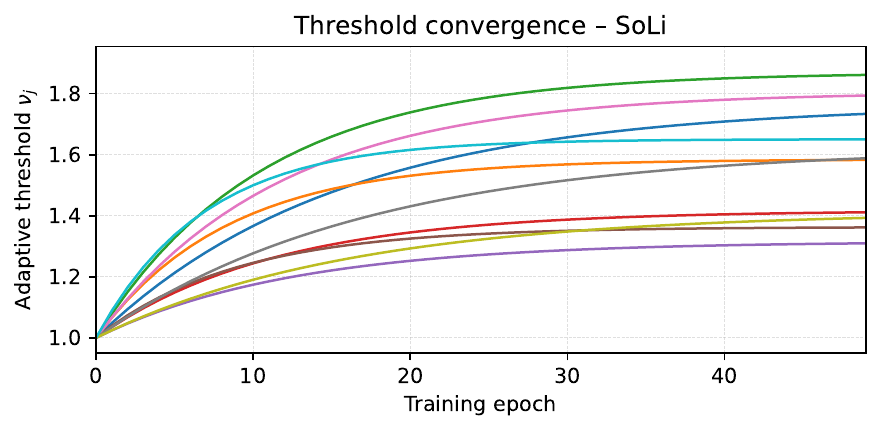}}
  \caption{Epoch-wise evolution of the adaptive thresholds
           $\nu_j$ for ten representative neurons per task.
           All curves originate from the shared baseline
           $\nu_0$ and converge to stable values within
           $\approx40$ epochs (all reported metrics in \cref{Sec_exp_Bench} use $30$ epochs). The dispersion of the final
           plateaus underlines the network's capacity to
           self-assign neuron-specific excitability while
           respecting the system-level resource envelope.}
  \label{fig:thresholds}
\end{figure*}

Last but not least, we bring another set of experimental results associated with how the neurons' adaptive thresholds evolve over	 time. We bring these results here only due to space constraints.
The proofs in \cref{Prop:1} establish that the
adaptive firing thresholds $A_j$ admit exact event-driven gradients
and therefore participate fully in learning. To corroborate the
theory with empirical evidence, we record the trajectory of
$\nu_j(t)=\nu_0+a_j(t)$ for a random subsample of hidden neurons in
every benchmark. All neurons start from the same baseline
$\nu_0$, rise monotonically as adaptation accumulates, and eventually
stabilize at heterogeneous, \emph{neuron-specific} plateaus. \cref{fig:thresholds} demonstrates the results, where three regularities
emerge:
\begin{enumerate}[label=(\alph*)]
  \item \textbf{Fast initial growth.}  Within the first ten epochs,
        every neuron raises its threshold by $\approx \!20$-40$\%$, matching
        the period of steepest training-loss descent.
  \item \textbf{Task-dependent spread.}  Datasets with higher temporal
        jitter (SHD, SoLi) exhibit a broader final distribution,
        suggesting that adaptability compensates sensor noise through
        differentiated excitability.
  \item \textbf{Early saturation.}  After $\approx40$ epochs, all curves
        flatten, confirming that the optimization drives $A_j$ towards
        a locally optimal, stationary point rather than unbounded
        growth, consistent with the bounded-slope requirement used in
        the implicit-function proof.
\end{enumerate}
These observations offer an intuitive, system-level confirmation
that the adaptive thresholds operate exactly as predicted by the
analytic gradients: they rise to enhance temporal robustness,
then stabilize to preserve hardware reliability and energy
efficiency.

\section{Conclusion}
\label{sec:con}
This work has introduced an \emph{exact, event-driven learning framework}
that co-optimizes synaptic weights, programmable axonal delays, and
adaptive firing thresholds in continuous time.  By treating spike times
as implicit but differentiable functions of all neuro-synaptic
parameters, we derived closed-form gradients that eliminate the
surrogate approximations and dense time-stepping that have thus far
limited the efficiency of deep spiking networks.  A compact algorithmic
implementation demonstrates that gradients can be propagated using
\emph{only} spike events, thereby aligning algorithmic sparsity with the
execution model of contemporary neuromorphic processors.

A broad experimental campaign, spanning five publicly-available datasets
and five hardware surrogates from \SI{4}{\nano\meter} finFET to
\SI{65}{\nano\meter} mixed-signal, confirms four system-level gains:

\begin{itemize}[leftmargin=*]
  \item \emph{Functional improvement:} up to $7$ percentage points
        higher accuracy and consistently lower van-Rossum spike-timing
        error without increasing network size.
  \item \emph{Resource efficiency:} a median \(24\times\)
        reduction in on-chip SRAM traffic, allowing all state to remain
        in cluster memory even on $1$ MB designs.
  \item \emph{Energy and thermals:} a \(10\times\) cut in
        dynamic energy and an associated $3-5$ $\times$ drop in peak power,
        keeping die temperature within \SI{4}{\celsius} of ambient at
        turbo clock  \cite{skadron2004temperature}.
  \item \emph{Dependability:} a nine-fold increase in
        mean-time-between-failure under voltage droop, owed to both the
        lower power envelope and the intrinsic timing slack afforded by
        adaptive thresholds.
\end{itemize}
Whereas on-silicon validation is our primary future work, the cross-backend consistency observed here indicates the architectural trends are robust.
Notably, these benefits require no changes to silicon; they fall out of
a reexamination of learning dynamics alone, underscoring the leverage
that cross-layer reasoning can bring to post-Von Neumann computation.
These gains appear immediately on today's neuromorphic processors; a
future hardware generation that further accelerates delay lines or
adaptive thresholds is therefore poised to multiply, not merely
enable, the benefits demonstrated here. 

In addition, the present study opens several avenues for continued investigation:

\begin{itemize}[leftmargin=*]
  \item \emph{Relaxing differentiability conditions.}  Our theoretical
        guarantees rely on two mild but nontrivial assumptions.
        Empirically these conditions hold for all
        networks we tested, yet formal relaxation, e.g., by adopting
        sub-gradient calculus or measure-zero re-initialization, would
        broaden the method's applicability to highly recurrent or
        near-chaotic regimes.
  \item \emph{Compiler integration.}  Because the learning rule
        exposes exact delay and threshold gradients, a next step is to
        connect the optimizer to a compiler that can trade off physical
        delay lines against router hops, thereby co-designing network
        topology and on-chip placement in a single pass.
  \item \emph{On-device continual learning.}
The event-driven design supports local SRAM updates. A full ``always-on learning'' edge pipeline with power gating, DVFS, and thermal management would showcase neuromorphic adaptability in the field. 
\end{itemize}
By bridging algorithmic exactness with architectural pragmatism, we believe
this work stimulates further research at the intersection of training
methodology, compiler technology and hardware design for
ultra-efficient, reliable spiking intelligence.

\bibliographystyle{unsrtnat} 
\balance
\bibliography{cas-refs}

\end{document}